\documentclass[sn-mathphys-num]{sn-jnl}

\usepackage{hyperref}
\usepackage{float}
\usepackage{verbatim} 
\usepackage{listings}%

\restylefloat{figure}
\restylefloat{table}

\usepackage{amsmath,amssymb,amsfonts}
\usepackage{array}
\usepackage{mdwmath}
\usepackage{mdwtab}
\usepackage{eqparbox}
\usepackage{url}
\usepackage[T1]{fontenc}
\usepackage{xcolor}

\usepackage{tcolorbox}
\usepackage[outline]{contour}
\usepackage{framed}
\usepackage{mdframed}
\usepackage{wrapfig}
\usepackage{enumerate}
\usepackage[vlined,linesnumbered,ruled,longend]{algorithm2e}
\usepackage{caption}
\usepackage{subcaption}

\usepackage{soul} 
\sethlcolor{yellow}

\usepackage{caption}
\usepackage{booktabs}
\usepackage{tabularx}
\usepackage{multicol}
\usepackage{multirow}
\usepackage{amssymb}
\usepackage{pifont}

\newtheorem{theorem}{Theorem}

\newtheorem{definition}[theorem]{Definition}

\newenvironment{proofof}[1]{\begin{trivlist} \item {\bf Proof #1:~~}}  {\qed\end{trivlist}}

\newcommand{\RIFconstruct}{{\textsc{RIFConstruction}}}
\newcommand{\scoring}{{\textsc{Scoring}}}
\newcommand{\combinescore}{{\textsc{CombineScores}}}
\newcommand{\forest}{\ensuremath{\mathcal{F}}}
\newcommand{\itree}{\ensuremath{\mathcal{T}}}
\newcommand{\rotationmatrix}{\ensuremath{\mathcal{R}}}

\begin{document}

\title[Detecting Anomalies Using Rotated Isolation Forest]{Detecting Anomalies Using Rotated Isolation Forest}


\author[a]{\fnm{Vahideh} \sur{Monemizadeh}}\email{v.monemizadeh@semnan.ac.ir}

\author*[a]{\fnm{Kourosh} \sur{Kiani}}\email{kourosh.kiani@semnan.ac.ir}

\affil[a]{\orgdiv{Electrical and Computer Engineering Department}, \orgname{Semnan University}, \orgaddress{\street{Sukan Park}, \city{Semnan}, \postcode{3513119111}, \state{Semnan}, \country{Iran}}}



\abstract{The Isolation Forest (iForest), proposed by Liu, Ting,  and Zhou at TKDE 2012, has become a prominent tool for unsupervised anomaly detection. 
However, recent research by Hariri, Kind, and Brunner, published in TKDE 2021, has revealed issues with iForest. They identified the presence of axis-aligned ghost clusters that can be misidentified as normal clusters, leading to biased anomaly scores and inaccurate predictions. 
In response, they developed the Extended Isolation Forest (EIF), which effectively solves these issues by eliminating the ghost clusters introduced by iForest. This enhancement results in improved consistency of anomaly scores and superior performance. We reveal a previously overlooked problem in the Extended Isolation Forest (EIF), showing that it is vulnerable to 
ghost inter-clusters between normal clusters of data points. 
In this paper, we introduce the Rotated Isolation Forest (RIF) algorithm which effectively addresses both the axis-aligned ghost clusters observed in iForest and the ghost inter-clusters seen in EIF.  RIF accomplishes this by randomly rotating the dataset (using \emph{random rotation matrices} and 
\emph{QR decomposition}) before feeding it into the iForest construction, thereby increasing dataset variation and eliminating ghost clusters.
Our experiments conclusively demonstrate that the RIF algorithm outperforms iForest and EIF, as evidenced by the results obtained from both synthetic datasets and real-world datasets.
}

\keywords{Anomaly Detection, Isolation Forest, Random Rotation.}


\maketitle

\section{Introduction}\label{sec1}

Anomaly detection refers to the process of identifying patterns or instances in data that deviate significantly from the norm or expected behavior. In other words, it involves finding data points that are rare, unexpected, or abnormal compared to the majority of the dataset. Anomalies can represent unusual events, errors, or potential threats, depending on the context of the application. The goal of anomaly detection is to pinpoint these unusual data points so that appropriate action can be taken, such as investigating potential fraud \cite{POURHABIBI2020113303,10.1007/978-981-16-5640-8_3}, detecting network intrusions \cite{10072303,10.1007/978-3-031-27409-1_123,stepanov2021detecting,chun2024random}, time series problems \cite{sorbo2023navigating,lu2023damp,zhang2021aurora}, contextual problems \cite{calikus2022wisdom,li2023explainable}, healthcare monitoring \cite{PACHAURI2015325,biedebach2023anomaly}, or monitoring equipment for signs of malfunction \cite{CHOI2022109147}.

\emph{Isolation Forest} (iForest), introduced by Liu, Ting, and Zhou \cite{DBLP:journals/tkdd/LiuTZ12} at TKDD'12 is a commonly used algorithm for detecting anomalies. 
Major commercial platforms such as Amazon Web Services (AWS)\footnote{https://docs.aws.amazon.com/sagemaker/latest/dg/randomcutforest.html}, Microsoft Azure\footnote{https://azure.microsoft.com/en-us/products/ai-services/ai-anomaly-detector}, and IBM AIOps\footnote{https://www.ibm.com/blog/anomaly-detection-machine-learning/} have incorporated iForest into their anomaly detection systems. 
The iForest algorithm works by isolating anomalies instead of studying normal data points. The main idea behind iForest is that anomalies are \emph{few} and \emph{different}. 
This assumption comprises two key ideas:
\begin{itemize}
    \item \emph{Majority Assumption:} The quantity of \emph{normal} data points significantly exceeds the count of \emph{anomalies}.
    \item \emph{Deviation Assumption:} The attribute values of \emph{anomalies} significantly differ from those of \emph{normal} data.
\end{itemize}

The construction of iForest works by randomly sub-sampling data, randomly selecting a feature and a random split value between the maximum and minimum values of the selected feature to isolate anomalies. This process is repeated recursively, forming a structure resembling a tree. Anomalies are expected to be isolated in fewer steps compared to normal data points, making them easier to detect. 
The iForest is recognized for its efficiency in handling high-dimensional data, scalability to large datasets, and resilience to outliers.

The partitioning strategy employed by iForest, which aligns with the axes, can create artifacts known as ghost clusters (See Figure~\ref{fig:ghost:oneblob:without:noise}), diminishing iForest's effectiveness in anomaly detection. These ghost clusters arise because data is split along random features using axis-aligned hyperplanes. Consequently, rectangular ghost clusters form, often aligning with normal clusters. This alignment leads to biased anomaly scores for true anomalies, resulting in misleading predictions.

The Extended Isolation Forest (EIF) \cite{DBLP:journals/tkde/HaririKB21}, introduced by Hariri, Kind, and Brunner at TKDE'21, represents an advancement over the original iForest algorithm. It aims to address the bias introduced by axis-aligned splits, which can result in inconsistencies in anomaly scores. EIF tackles this issue by employing splits based on randomly chosen directions, represented by hyperplanes with random slopes that are not aligned with the axes. In contrast to iForest, which uses axis-aligned hyperplanes, this approach helps eliminate the ghost clusters created by iForest, leading to improved consistency in anomaly scores.

In this paper, we carefully examine the iForest and EIF and reveal a previously overlooked problem in the Extended Isolation Forest (EIF), showing that it is vulnerable to 
ghost inter-clusters between normal clusters of data points (See Figure~\ref{fig:ghost:two:blobs}).
This phenomenon in EIF results in low anomaly scores for truly anomalous points and therefore, inaccurate predictions. 
This adds to the already complicated nature of EIF, especially when it tries to split points in many different directions in spaces with lots of dimensions.

In this paper, we introduce the Rotated Isolation Forest (RIF) algorithm which effectively addresses both the axis-aligned ghost clusters observed in iForest and the ghost inter-clusters seen in EIF.  RIF accomplishes this by randomly rotating the dataset before feeding it into the iForest construction. 
Consequently, the RIF effectively eliminates the ghost clusters created by iForest and EIF and 
enhances the consistency of anomaly scores, resulting in improved predictions as we will see in the results section. 
In summary, RIF offers the following advantages over iForest and EIF: 
\begin{enumerate}
    \item EIF uses splits in various directions, while RIF utilizes a single random rotation of the dataset.This, in turn, decreases the number of random bits utilized by RIF.
    \item The construction of RIF is simpler than EIF because it avoids generating hyperplanes with random slopes for high-dimensional spaces.
    \item RIF yields more consistent anomaly scores and enhances prediction accuracy by eliminating spurious ghost regions.
\end{enumerate}

It is worth mentioning that the idea of randomly rotating datasets before applying the iForest construction was also explored in \cite{DBLP:journals/tkde/HaririKB21}. However, the authors faced various challenges with this approach, leading them to adopt a more sophisticated partitioning method using hyperplanes with random slopes, as proposed in EIF. \cite{DBLP:journals/tkde/HaririKB21}. Specifically, they outlined the following issues they encountered:\emph{"This approach can become cumbersome to apply
especially with large datasets and higher dimensions. The rotation is not obvious in higher dimensions
than 2-D. For each tree we can pick a random axis in the space and perform planar rotation around that
axis, but there are many other choices that can be made, which might result in inconsistencies among
different runs"}.  

In this paper, we initially present a simple yet innovative mechanism for randomly rotating datasets of high dimensions using \emph{random rotation matrices} and 
\emph{QR decomposition}. We show in Subsection~\ref{sec:issues} that this combination resolves the challenges faced by the authors of \cite{DBLP:journals/tkde/HaririKB21}. 
Subsequently, we highlight the superior efficiency of the RIF algorithm compared to its counterparts by addressing and overcoming the challenges encountered with iForest and EIF.

\begin{figure}[h]
    \centering
    \includegraphics[width=0.9\linewidth]{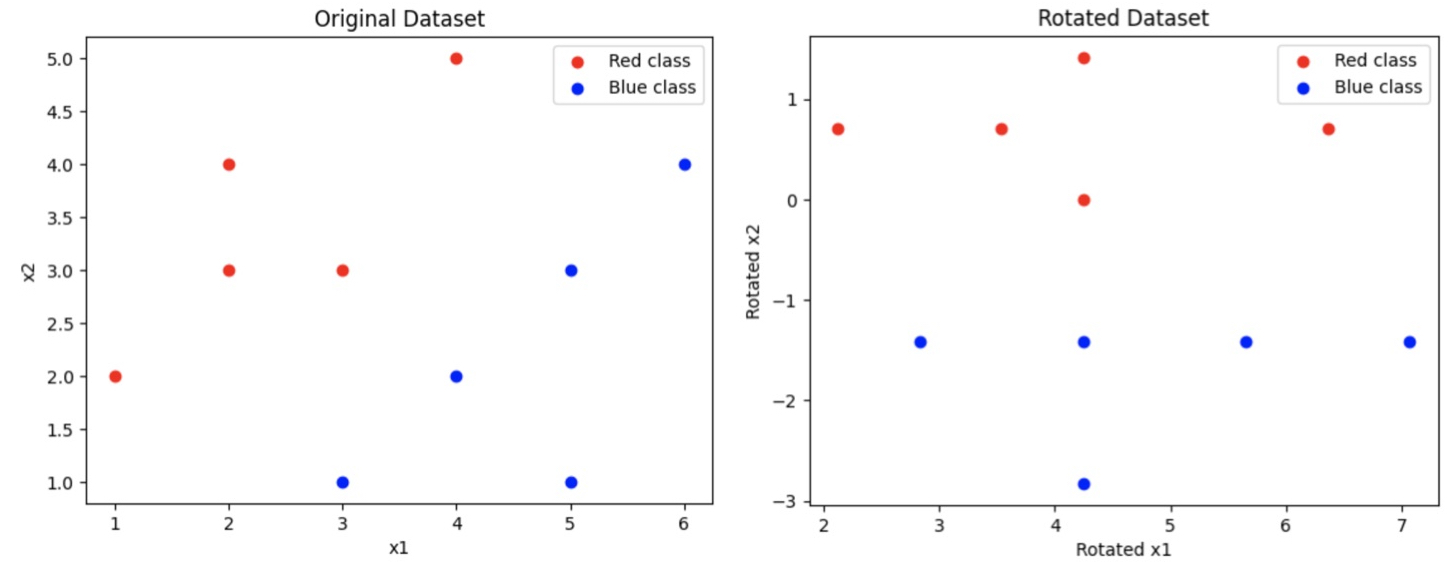}
    \caption{Rotation improves separability.}
    \label{fig:rotation:help}
\end{figure}

Finally, we provide rationale for our decision to randomly rotate datasets prior to feeding them into the construction algorithm for iForest. 
Rotations frequently enhance separability in classification and anomaly detection tasks. To exemplify, let's examine a two-dimensional dataset characterized by features $x1$ and $x2$, with two classes denoted by red and blue points. In Figure~\ref{fig:rotation:help}, we visualize the original dataset, where a basic vertical or horizontal (axis-aligned) separator encounters difficulty in effectively distinguishing between the two classes.

However, by introducing clockwise or counterclockwise rotations to the feature space, we aim to amplify class separability. As the rotation angle nears $45$ degrees ($\pi/4$ radians), we observe a noticeable enhancement in separability between the classes. This enhancement holds potential to significantly improve classification performance when employing classifiers within the rotated feature space. Consequently, both vertical and horizontal separators can accurately discern between blue and red classes.

Nonetheless, the optimal rotation angle for improving dataset separability remains unknown. To address this, we employ an ensemble of random rotations on the dataset to enhance the separability criteria. Through this approach, we aim to identify the rotations that yield the most favorable separability outcomes.

\textbf{Outline:} 
The rest of the paper is organized as follows: 
In Section $2$, we first explain the state of the art results known for anomaly detection in Euclidean spaces. This section is followed by Section $3$, Preliminaries, where we define a few terminologies. Section $4$ provides an insightful review of the original iForest and EIF approaches for anomaly detection. In Section $5$, we explain the process of generating a random rotation matrix. Section $6$ presents our Rotated Isolated Forest (RIF). Extensive experimental results



\section{Related work}
Anomaly detection algorithms can generally be grouped into several classes. 
Additionally, refer to the survey conducted by Samariya and Thakkar \cite{samariya2023comprehensive} for a more in-depth exploration of the latest advancements in anomaly detection algorithms designed for Euclidean spaces.

\paragraph{Classification-based approaches}
These methods utilize training data to learn parameters for various underlying classification models. Examples include neural networks \cite{chalapathy2018anomaly}, Bayesian networks \cite{mascaro2014anomaly}, support vector machines \cite{li2003improving}, and rule-based models \cite{duffield2009rule}. While these models can offer effective and efficient detection performance, the availability or relevance of high-quality training data may be limited.

\paragraph{Distance-based approaches}
    These methods rely on a distance metric between data points. An advantage of these algorithms is their unsupervised nature, eliminating the need for labeled data. This class can be further subdivided as follows:
    \begin{itemize}
        \item \emph{Nearest-neighbors methods:} These techniques evaluate the distance of a point to its $k$-th nearest neighbor or $k$ nearest neighbors, or the distance within some other local neighborhood, to assess the anomaly level of a point. For instance, the Local Outlier Factor \cite{breunig2000lof} calculates a relative density value for a point using its $k$-nearest neighbors (similar to concepts in DBSCAN \cite{ester1996density}) to address density variations across different clusters of normal data.
        \item \emph{Clustering methods:} These approaches aim to cluster the data, identifying points that do not belong to any cluster as anomalies. Commonly used clustering algorithms include $k$-means \cite{munz2007traffic} and DBSCAN \cite{ccelik2011anomaly}.
    \end{itemize}
    Both subclasses share similarities, such as assuming the existence of local neighborhoods in the data. The performance of these methods also depends on the choice of distance measure.

\paragraph{Statistical-based approaches}
    This category of algorithms can be further divided into the following subcategories:
    \begin{itemize}
        \item \emph{Parametric techniques:} Methods in this subclass model datasets as samples drawn from an underlying statistical distribution. Examples include the use of Gaussian \cite{laxhammar2008anomaly} or regression \cite{salem2014anomaly} models.
        \item \emph{Nonparametric techniques:} Methods in this subclass do not assume a specific statistical distribution. Examples include histograms \cite{goldstein2012histogram} and kernel functions \cite{kwon2005kernel}.
    \end{itemize}
    These categories offer robust methods that can provide statistical guarantees about their results. However, describing the underlying data distribution and understanding the interaction between various features can be challenging in practice.

\paragraph{Ensemble-based approaches \cite{fontugne2010mawilab}}
    This category of methods combines the outcomes of multiple anomaly detectors along with a consensus mechanism to determine a final anomaly labeling. By leveraging diverse anomaly detectors that complement each other and are not affected by the same limitations, these methods prioritize robustness over runtime complexity.

\paragraph{Subspace-based approaches}
    These methods conduct anomaly detection on various reduced subspaces of the full feature space. For instance, one approach involves utilizing random Gaussian projection to obtain subspaces and subsequently analyzing them \cite{ding2013compressed}. While these methods are reputed for their efficacy in uncovering "hidden" anomalies, exploring numerous subspaces can entail high computational expenses, potentially resulting in unnecessary computational overhead.

\paragraph{Isolation-based approaches}
    Isolation-based anomaly detection algorithms aim to separate anomalies from normal data by making early cuts. Empirical studies cited in  \cite{emmott2013systematic} have shown that isolation-based algorithms using randomization techniques outperform other anomaly detection methods, such as distance-based \cite{breunig2000lof, tang2002enhancing, papadimitriou2003loci, jin2006ranking, kriegel2009loop}, FUZZY C-MEANS Based Extended Isolation Forest \cite{10.1007/978-3-031-26384-2_35}, Generalized Isolation Forest \cite{10.1007/978-3-031-57853-3_30} and density-based \cite{chandola2009anomaly, pn2005introduction, ester1996density} algorithms. The Isolation Forest, also known as iForest \cite{DBLP:journals/tkdd/LiuTZ12}, is considered the foundational algorithm in this category. Recent advancements have led to variations of this algorithm, including Extended Isolation Forest (EIF) \cite{DBLP:journals/tkde/HaririKB21} and Robust Random Cut Forest (RRCF) \cite{guha2016robust}. Despite demonstrating impressive performance and quick execution times, these methods still have limitations in terms of accuracy and scalability. These limitations will be further explored later in this paper.
    
%
%
%
%
%
%
%
%
%
%
%


\section{Preliminaries}
An anomaly detection algorithm functions akin to a binary classifier, aiming to differentiate between normal and anomalous instances within a dataset $\mathcal{D}$ containing $n$ entries. Here, $n_1$ entries are labeled as anomalies, with $n_0$ representing normal data, determined by ground truth labeling. The \emph{contamination} of the dataset, denoted by $c = \frac{n_1}{n}$, is indicative of the ratio of anomalies to the total number of entries. The algorithm's objective is to classify each entry as either $True$ if it shows an anomaly or $False$ if it denotes a normal instance. Typically, assessing classifier performance involves comparing its predictions to the ground truth labeling of $\mathcal{D}$.

The anomaly detection algorithms discussed in this paper produce continuous anomaly scores for dataset entries, normalized between $0$ and $1$. 
A higher anomaly score indicates a higher degree of anomaly. To convert these scores into binary labels, we pick the $n_1 = c \times n$ entries 
with the highest anomaly scores as $True$, while the remaining entries are labeled $False$. 

We measure the performance of anomaly detection algorithms using the the \emph{Area Under the Curve (AUC)}.  The AUC score is a widely used metric in machine learning, particularly in binary classification tasks. It quantifies the performance of a classification model across different thresholds for distinguishing between the positive and negative classes. The AUC represents the probability that the model will assign a higher score to a randomly chosen positive instance compared to a randomly chosen negative instance. 

In essence, the AUC measures the ability of the model to rank positive instances higher than negative instances, regardless of the specific threshold chosen. A higher AUC score indicates better discrimination performance, with a score of $1$ representing perfect classification and a score of $0.5$ indicating random guessing. AUC is favored over other metrics like accuracy in situations with imbalanced datasets or when the true positive rate and false positive rate need to be evaluated across various decision thresholds. Overall, AUC provides a comprehensive assessment of a classification model's predictive power and robustness.


\section{iForest and Extended iForest}
The Isolation Forest (iForest) method was introduced by Liu, Ting, and Zhou \cite{DBLP:journals/tkdd/LiuTZ12}, who identified a key limitation in prior anomaly detection approaches. 
Traditionally, such methods (e.g., clustering-based methods) primarily focused on building profiles of normal data, rather than directly targeting anomalies. 
In contrast, iForest places emphasis on directly detecting anomalies. 
We next briefly explain the construction of iForest.

\subsection{Isolation Forest (iForest) Construction} 
Consider a point set $P \subset \mathbb{R}^d$ comprising $n$ points in a $d$-dimensional Euclidean space $\mathbb{R}^d$.
The iForest algorithm constructs an iForest $\forest$ which is an ensemble of $t$ iTrees. 
Each iTree $\itree_i$ for $i \in \{1, 2, \cdots, t\}$ samples independently and uniformly at random 
a subset $X_i \subseteq P$ of size $\psi$ uniformly at random. 
Liu, Ting, and Zhou \cite{DBLP:journals/tkdd/LiuTZ12} empirically demonstrate that optimal parameter choices for $t$ and $\psi$ are $\psi = 256$ and $t = 100$.

An iTree is constructed as a recursive binary partition tree on its subsample, achieving splits by iteratively dividing among a uniformly random dimension at a random value. 
This random value is uniformly generated within the range of the current subset along the split dimension. The splitting process continuous until either only one point 
remains (\emph{isolated point}) or a predefined depth limit is reached (defaul $\lceil \log_2 \psi \rceil$). 

The path length to the leaf containing a point in an iTree indicates the number of splits required to isolate it. This serves as a metric for the anomaly level of a point; a shallower depth suggests a higher anomaly level (i.e., easier isolation), while a deeper depth indicates that the point is more closely connected to other points and is more likely to be a normal point. When a leaf $\ell$ reaches the depth limit, a penalty score is introduced to compensate for the absence of further splits. The subtree of the iTree rooted at $\ell$ is assumed to follow a Binary Search Tree (BST) structure for the points of leaf $\ell$. However, we do not construct this BST. Instead, we apply a penalty term based on the average search path length of a BST to all points in leaf $\ell$. The primary reason for this approach is that in iForest, the main objective is to segregate anomalies from normal data, rather than distinguishing individual normal data points from each other.
We next explain the scoring functions for every iTree $\itree_i$ 
and for the iForest $\forest$.

\paragraph{Scoring function for iTree}
Let $dep$ denote the depth limit set for iTrees in iForest $\forest$, and $x \in P$ be an arbitrary point. The anomaly score of $x$ for iTree $\itree_i$, denoted as $\mathcal{A}(x, \itree_i)$, is computed as follows:
\begin{itemize}
    \item If the leaf $\ell_x$ in iTree $\itree_i$ containing $x$ is at a level less than $dep$, then $\mathcal{A}(x,\itree_i) = Level (\ell_x)$, where $Level (\ell_x)$ represents the level of leaf $\ell_x$ in iTree $\itree_i$.
    \item If the leaf $\ell_x$ in iTree $\itree_i$ containing $x$ is at level $dep$, then $\mathcal{A}(x,\itree_i) = dep + c(n)$, where $c(n)\footnote{$H(i)$ is the $i$-th harmonic number, approximated as $\ln(i) + \gamma$, with $\gamma \approx 0.5772156649$ representing the Euler–Mascheroni constant.} = 2H(n-1) - \frac{2(n-1)}{n}$ 
    denotes the average path length of an unsuccessful search in a Binary Search Tree (BST) of size $n$ and $n$ is the number of points inserted into the leaf. 
\end{itemize}

\paragraph{Scoring function for iForest}
Once an iForest is constructed, it becomes capable of computing the anomaly score for every point $x \in P$ and in general, for any $x \in \mathbb{R}^d$. 
The anomaly score of point $x$ is evaluated in every iTree $\itree_i$ within the iForest $\forest$ and the average score of all scores returned by all iTrees is computed. 
Define $H(x) = E[\mathcal{A}(x, T_i)] = \sum_{j=1}^t \mathcal{A}(x,T_i) / t$ as the average anomaly score of $x$ across all iTrees within the iForest $F$.
The anomaly score of $x$ for iForest $F$, denoted as $\mathcal{A}(x,F)$, is given by 
$\mathcal{A}(x,F) = 2^{- H(x) / c(\psi)}$. 
The final score is normalized to ensure that a score of $0$ indicates a normal point, while a score of $1$ presents an anomalous point.  
It is important to note that while every point $x $ is processed through the iTrees for scoring, they are not stored in iTrees.


\subsection{Ghost cluster phenomenon} 
A key question about iForest's output is understanding its weaknesses. 
It's valuable to look at datasets where normal and anomalous data are clearly separated to see how well iForest performs. 
To this end, Hariri, Carrasco Kind, and Brunner \cite{DBLP:journals/tkde/HaririKB21} have examined datasets with points generated from one or
 two spherical Gaussian distributions to evaluate the performance of iForest in these scenarios.
They identified a significant phenomenon in the context of iForest construction. 
In particular, they observed that iForest algorithm generates what they termed as \emph{ghost clusters}, leading to an increased incidence of false negatives.


\begin{figure}[h]
    \centering
    \includegraphics[width=0.8\linewidth]{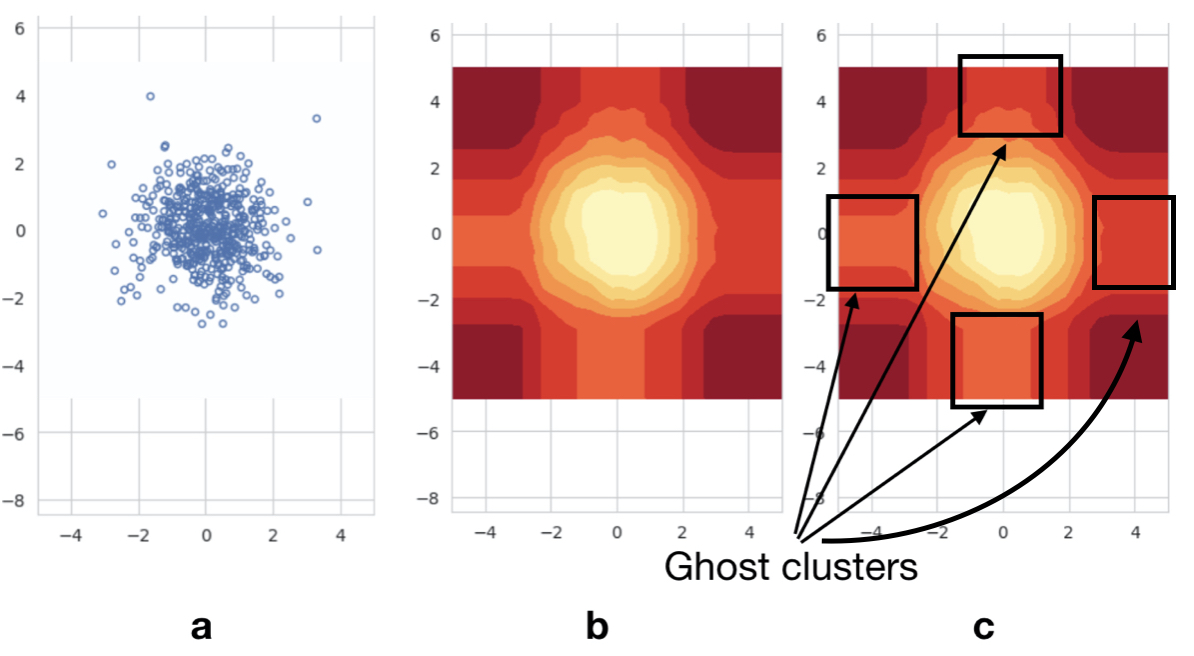}
    \caption{The phenomenon of ghost clusters is observed in the output of iForest for a single Gaussian distribution. }
    \label{fig:ghost:oneblob:without:noise}
\end{figure}


In Figure~\ref{fig:ghost:oneblob:without:noise}, we underscore the phenomenon of ghost clusters apparent in the iForest algorithm's output. For this experimental setup, we generate a point set following a spherical Gaussian distribution $X \approx N(\mu=0, \sigma^2=1)$, from which we sample a set of $1000$ points. Sub-figure (a) visually depicts the distribution of these sampled points. In Sub-figure (b), we present a heatmap illustrating the output of the iForest algorithm on this dataset. In this heatmap, lighter areas correspond to normal points, while darker red regions signify areas with higher anomalous scores. Sub-figure (c) specifically highlights regions identified as ghost clusters.

The key observation here is that, given the dataset's generation from a spherical Gaussian distribution, the lighter areas depicted in the figure should ideally resemble an annulus. However, contrary to this expectation, we observe lighter cross-shaped regions incorrectly classified as normal areas leading to false negatives. This misclassification poses a significant challenge, as any anomaly point situated within these areas is erroneously considered normal. Such instances of misidentification compromise the effectiveness of the iForest algorithm in accurately distinguishing anomalies, underscoring the need for a more refined approach in handling the generation of ghost clusters.

The observation in \cite{DBLP:journals/tkde/HaririKB21} underscores the need to refine the iForest algorithm, specifically targeting the challenge of ghost clusters. Such refinement efforts aim to improve the algorithm's accuracy in anomaly detection, reducing the incidence of false negatives in anomaly detection scenarios.

\subsection{Extended Isolation Forest (EIF) Construction}
In response to the ghost cluster dilemma, Hariri, Carrasco Kind, and Brunner \cite{DBLP:journals/tkde/HaririKB21} introduced 
an enhanced version of the iForest algorithm known as the \emph{Extended Isolation Forest (EIF)}. The EIF incorporates two distinct enhancements:

\begin{enumerate}
    \item \emph{Input rotations:} 
    Each iTree $T_i$ undergoes a random rotation of the point set before we use the point set to construct the iTree. 
    The authors in \cite{DBLP:journals/tkde/HaririKB21} 
    left it as open problem to develop input random rotations in high dimensions and to study the performance of 
    random rotations in high dimensions.
    
    \item \emph{Intermediate rotations:} 
    In iForest, during any recursive step,  a randomly axis-aligned hyperplane is selected to split the points within the corresponding region. 
    In EIF, instead, a randomly \textbf{rotated} hyperplane (i.e., not necessarily axis-aligned) is chosen to split the points within that region.
One drawback of random intermediate rotations is its complexity in higher dimensions, although they were able to achieve this through sophisticated techniques.
\end{enumerate}

\subsection{Ghost clusters using iForest, EIF, and RIF}
In this section, we focus on tackling the ghost cluster problem and checking how well iForest, EIF, and our RIF algorithms work. To understand how these algorithms perform, we use datasets generated from one or two spherical Gaussian distributions, with anomaly points placed in specific spots. Our goal is for these algorithms to find these added anomaly points effectively. We evaluate their performance using a heatmap created from a fine grid placed on the distribution space, along with the AUC score given by each algorithm.


\begin{figure}[h]
    \centering
    \includegraphics[width=0.7\linewidth]{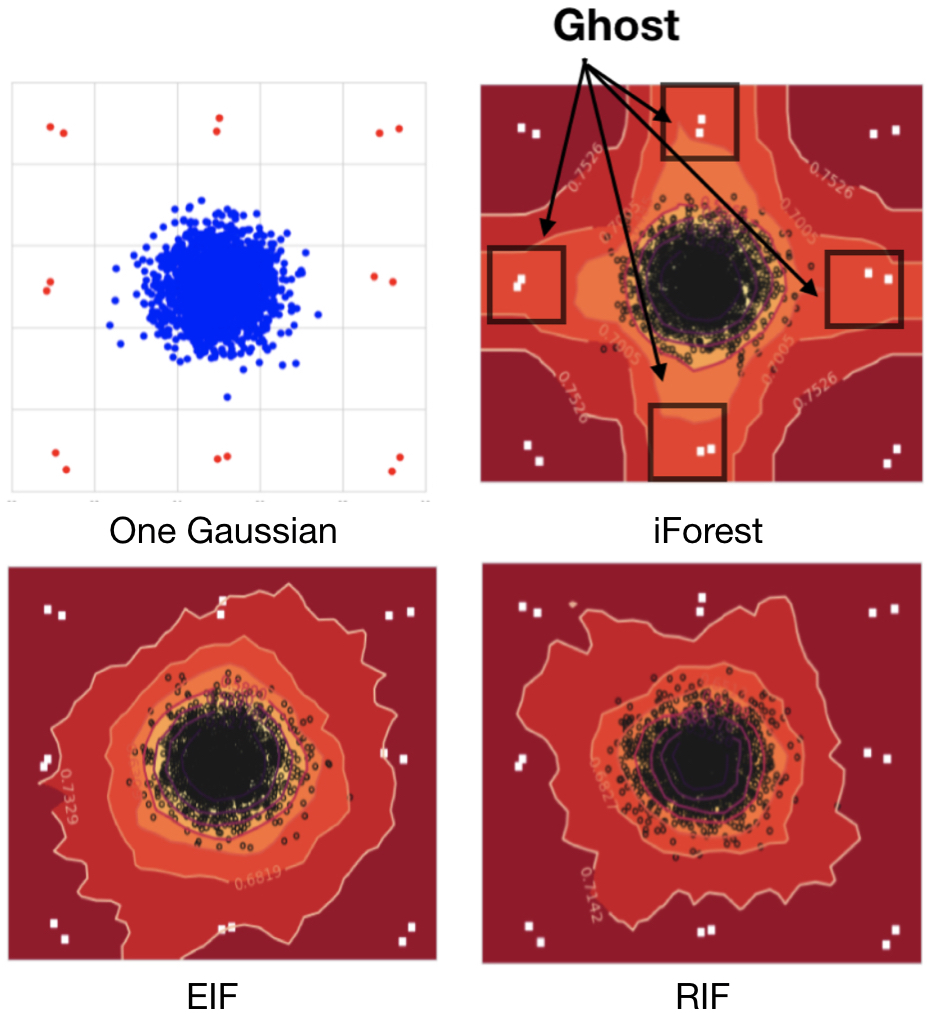}
    \caption{Ghost cluster phenomenon for one Gaussian distribution. }
    \label{fig:ghost:oneblob}
\end{figure}


We start by making a single round shape in the space $[0,1]^2$. This shape is in the middle at $\mu=(0.5,0.5)$ and has a spread of $\sigma=0.07$, making $2000$ points.

Next, we add anomaly points. We put $16$ anomaly points at each corner of the space $[0,1]^2$ and also to the north, south, east, and west. So, there are two anomaly points at each corner (top-right, bottom-right, bottom-left, and top-left), and two each to the north, south, east, and west. This gives us the dataset shown in the left part of Figure~\ref{fig:ghost:oneblob}.

We then do tests using iForest, EIF, and RIF on this dataset. Since we added $16$ anomaly points to a dataset of $2000$ normal points from a single round shape, the anomalous ratio in the data is about $\frac{16}{2016} \approx 0.008$.

For each method, we use $100$ trees and a sample size of $256$. We split the space $[0,1]^2$ into a $30 \times 30$ grid, which gives us $900$ points. Then, we use iForest, EIF, and RIF to calculate the anomaly score for these $900$ points. The heatmaps in Figure~\ref{fig:ghost:oneblob} show these results. 
Darker colors mean higher anomalous scores. Black points are normal, and white points are anomalous.

In the heatmap from iForest, we see ghost clusters at the top, right, bottom, and left. But in the heatmaps from EIF and RIF, there are no ghost clusters. The AUC scores are $0.87$ for iForest, and $0.99$ for both EIF and RIF. Thus, for this experiment, EIF and RIF outperform iForest.

\begin{figure}[h]
    \centering
    \includegraphics[width=0.7\linewidth]{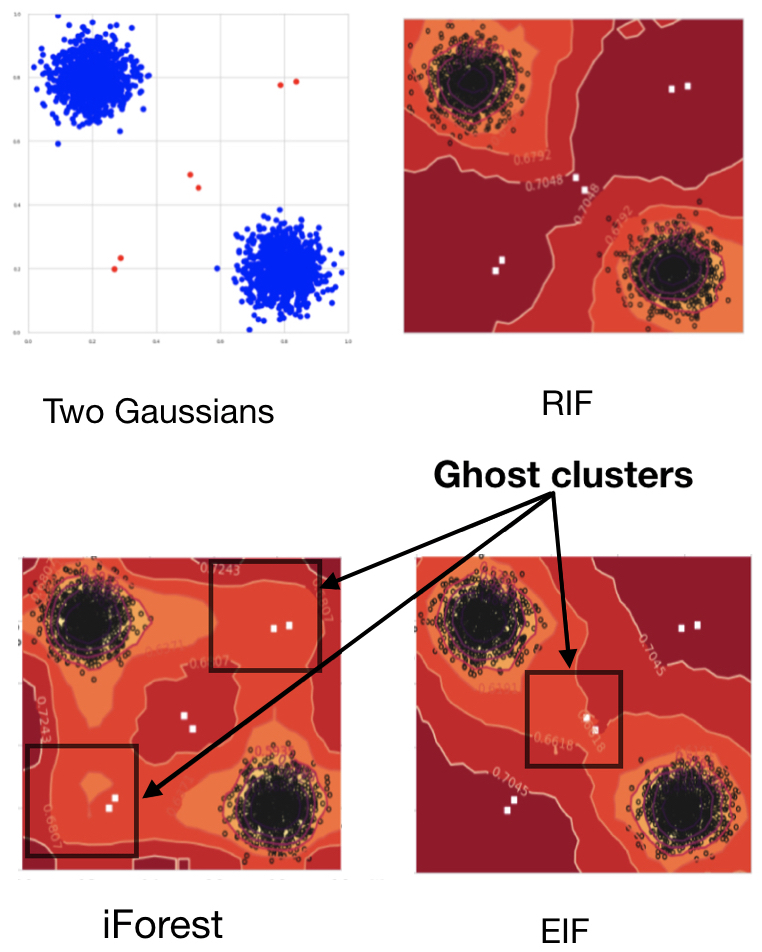}
    \caption{Ghost clusters phenomenon for a two Gaussian distributions. }
    \label{fig:ghost:two:blobs}
\end{figure}


Next, we look at two spherical Gaussians with centers at $(0.8, 0.2)$ and $(0.2, 0.8)$, each with a spread of $0.06$. We mix them equally to make a dataset of $2000$ points. 
We add $6$ anomaly points: two near $(0.8, 0.8)$ (top-right), two near $(0.25, 0.25)$ (bottom-left), and two in the middle at $(0.5, 0.5)$. These anomaly points are white in the left part of Figure~\ref{fig:ghost:two:blobs}, while normal points are blue. With $6$ anomaly points and $2000$ normal points, anomalies make up approximately $0.003$ of the dataset.

In the iForest heatmap, we see ghost clusters at the top-right and bottom-left. For EIF, the ghost clusters are between the two spherical Gaussians. But in the RIF heatmap, there are no ghost clusters. 
This essentially predicts that RIF should a better AUC score than iForest and EIF. 
Indeed, our experiments confirms this prediction and show that the AUC score of  RIF is $0.99$, while iForest scores $0.66$, and EIF has the AUC score of $0.83$.


\begin{figure}[h]
    \centering
    \includegraphics[width=\linewidth]{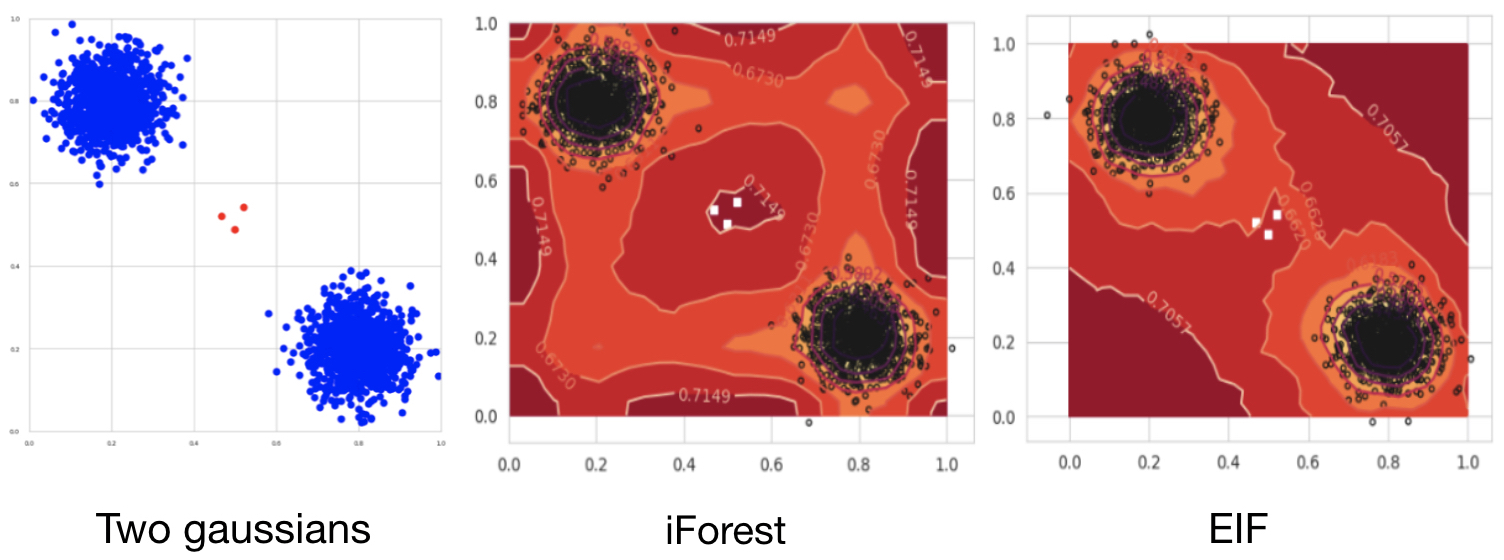}
    \caption{Two gaussian distributions with three anomaly points between two gaussian clusters.
    Contamination is $0.0045$ and the AUC score for iForest is $1.0$. However, the AUC score for EIF is $0.83$}
    \label{fig:ghost:if:vs:eif}
\end{figure}



Our latest experiment reveals an interesting finding that challenges conventional wisdom. 
In certain situations, iForest performs better than RIF, contrary to what was previously thought.  
To understand when iForest outperforms EIF, we repeat a similar experiment to the previous one. 
However, this time, we only place three anomaly points near the center of the two Gaussian clusters, specifically around the coordinates $(0.5, 0.5)$. Afterward, we run our algorithms, focusing only on iForest and EIF.

The results show that iForest successfully identifies all three anomaly points, while EIF only detects two out of three. 
Looking closer at the heatmaps generated by iForest and EIF, we observe a distinct darkened area in the iForest heatmap 
that is absent in the EIF heatmap. As a result, the ghost cluster in the middle of the EIF heatmap becomes more conspicuous. 
It is  important to note that ghost clusters are still visible at the corners of the iForest heatmap, 
indicating a potential decrease in performance when anomaly points are in those areas. However, when anomaly points are concentrated in the middle, iForest performs better than EIF.


\section{Random rotations in high dimensions}

Let us consider a $d \times d$ matrix $A$ which is indeed, a square matrix of order $d$. 
An \emph{orthogonal matrix} $A$ is a matrix that satisfies $A A^T = I$, where $A^T$ is the transpose of the matrix $A$ and $I$ is the identity matrix. 

\begin{definition}[Rotation matrix]
A rotation matrix $\mathcal{R}$ is a real-valued $d \times d$ orthogonal square matrix with a unit determinant. In other words, it satisfies the conditions:
\[
	\mathcal{R}^T = \mathcal{R}^{-1} \quad \text{ and } \quad |\mathcal{R}| = 1 \enspace ,
\]
where $\mathcal{R}^{-1}$ is  the inverse of the matrix $\mathcal{R}$, i.e., $\mathcal{R}\mathcal{R}^{-1} = \mathcal{R}^{-1}\mathcal{R} = I$. 

\end{definition}

The group of orthogonal matrices 
in dimension $d$, denoted by $O(d)$ (see Diaconis and Shahshahani \cite{Diaconis_Shahshahani_1987}), 
encompasses transformations that preserve distances in a $d$-dimensional Euclidean space while fixing a point. 
This group operation involves composing transformations. 
This orthogonal group consists of two connected components. 
\begin{itemize}
\item \textbf{Determinant $1$:} 
This component is known as the \emph{special orthogonal group}, denoted by $SO(d)$ and contains all orthogonal matrices with a determinant of $1$. 
This subgroup, also known as the \emph{rotation group}, extends the concept of rotations in dimensions $2$ and $3$, 
where its elements represent typical rotations around a point (in dimension $2$) or a line (in dimension $3$). 
\item \textbf{Determinant $-1$:}
The other component comprises all orthogonal matrices with a determinant of $-1$. However, this component does not form a group because the product of any two of its elements yields a determinant of $1$, rendering it incompatible with the component's determinant of $-1$.
\end{itemize}

We generate a $d \times d$ \emph{random rotation matrix} $Q$ in two steps:

First, we generate a \emph{random matrix} $A$ in \(d\)-dimensional space by sampling each entry independently from a standard normal (Gaussian) distribution with zero mean and unit variance. This means that each entry \(a_{ij}\) of the matrix is drawn from a random variable \(X\) that follows the normal distribution \(\mathcal{N}(0, 1)\), where the probability density function of \(X\) is given by 
$f(x) = \frac{1}{\sqrt{2\pi}} e^{-\frac{x^2}{2}}$. This process ensures that each entry of the matrix is randomly and independently chosen, providing a way to generate 
random matrices suitable for various applications such as dimensionality reduction \cite{van2009dimensionality}, random projections \cite{achlioptas2001database}, and statistical modeling \cite{konishi2008information}. 

Next, once we have the random matrix \(A\), we can perform QR decomposition \cite{householder1958unitary} to factorize it into an orthogonal matrix \(Q\) and an upper triangular matrix \(R\). 
The QR decomposition of \(A\) is given by $A = QR$, 
where \(Q\) is an \(d \times d\) orthogonal matrix and \(R\) is an \(d \times d\) upper triangular matrix. 
This decomposition is useful for various numerical algorithms, including solving linear equations \cite{waseem2013decomposition}, least squares regression \cite{harville1998matrix}, and eigenvalue computations \cite{abdi2007eigen}.
The matrix \(Q\) is a random rotation matrix in \(d\)-dimensional space. 

In summary, the process of generating a random rotation matrix involves the following steps:

\begin{enumerate}
    \item \textbf{Generate a random matrix (\(A\)):} Start by generating a random matrix \(A\) of size \(d \times d\) with elements drawn from a  normal probability distribution with mean \(0\) and variance \(1\).
    
    \item \textbf{Perform QR decomposition:} Perform QR decomposition on the random matrix \(A\) to obtain an orthogonal matrix \(Q\) and an upper triangular matrix \(R\), such that \(A = QR\).
    
\end{enumerate}

By following these steps, we obtain a random rotation matrix \(Q\).


\section{Rotated isolation forest in $d$-dimensional Euclidean space}
\begin{figure}[ht]
    \centering
    \includegraphics[width=0.9\linewidth]{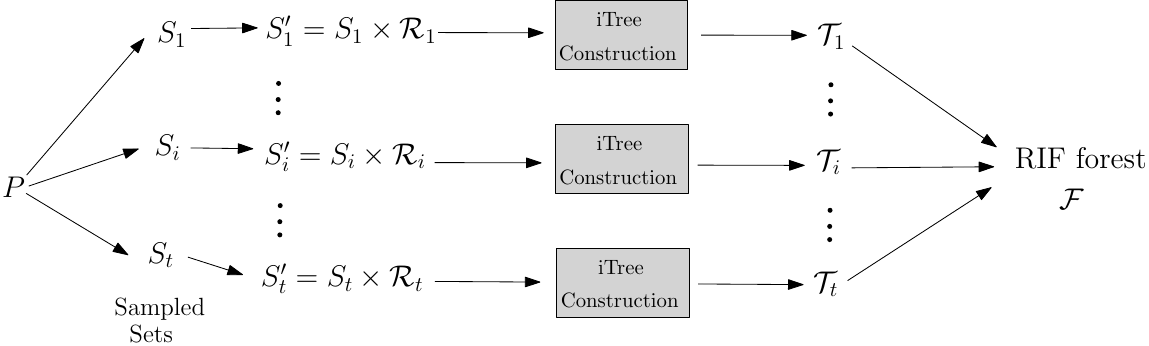}
    \caption{The pseudocode for constructing a RIF.}
    \label{fig:rif:construction}
\end{figure}

\begin{figure}[ht]
    \centering
    \includegraphics[width=0.9\linewidth]{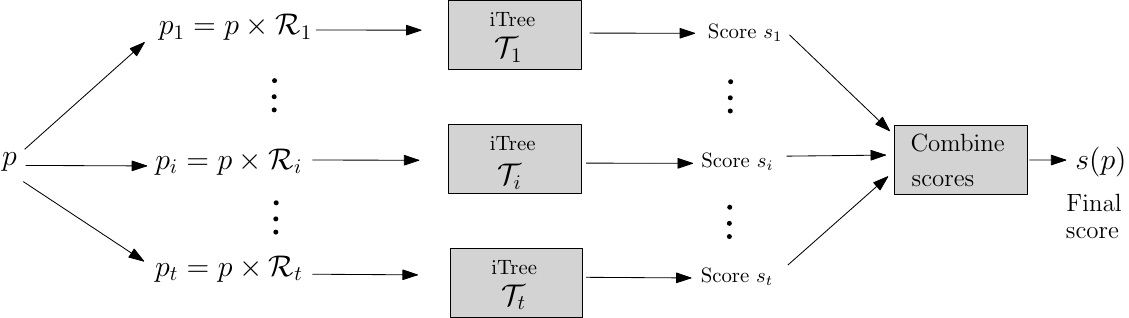}
    \caption{Computing the anomaly score for an arbitrary point $p$.}
    \label{fig:rif:anomaly:score}
\end{figure}

In this section, we explore our primary anomaly detection algorithm, known as \emph{Rotation Isolation Forest (RIF)}. 
This algorithm is comprised of two primary tasks: 
Firstly, we have the subroutine \RIFconstruct{}, outlined in Algorithm~\ref{alg:rif:construction} and Figure~\ref{fig:rif:construction}. This subroutine is responsible for constructing a RIF from a given point set $P$. 
Secondly, we introduce the subroutine \scoring{}, detailed in Algorithm~\ref{alg:rif:scoring} and Figure~\ref{fig:rif:anomaly:score}. 
This subroutine is designed to compute the anomaly score of a point $p$, given a RIF $\forest$ and an arbitrary point $p$.


\begin{algorithm}[ht]
 \caption{{\RIFconstruct}}
\label{alg:rif:construction}
 \KwData{A set $P$ of $n$ points in $d$-dimensional Euclidean space $\mathbb{R}^d$, sample size $s$, and a number $t$ of trees}
 \KwResult{A RIF forest $\forest$}
Compute $t$ random rotation matrices $\rotationmatrix_1,\cdots,\rotationmatrix_t$\;
\For{$i=1$ to $t$}{
	Let $S_i$ be a set of $s$ samples drawn uniformly at random from point set $P$\;
	Let $S'_i = S_i\times \rotationmatrix_i$ be the rotation of sampled set $S_i$ using rotation matrix $\rotationmatrix_i$\;
	Construct iTree $\itree_i$ for sampled set $S'_i$\;
}
Return RIF forest $\forest = \cup_{i=1}^t (\itree_i, \rotationmatrix_i)$, where for every iTree  $\itree_i$, we also return its corresponding 
random rotation matrices $\rotationmatrix_i$
\end{algorithm}

The input for \RIFconstruct{} consists of a set $P \subseteq \mathbb{R}^d$ containing $n$ points, along with two parameters: $s$, representing the sample size, and $t$, indicating the number of trees to be constructed in the forest $\forest$. For each tree $\itree_i$ where $i \in [t]$, we initially compute a random rotation matrix $\rotationmatrix_i$. Subsequently, we sample a set $S_i$ uniformly at random from the point set $P$ and compute its rotated version $S'_i$ using the rotation matrix $\rotationmatrix_i$. This rotated set $S'_i$ serves as input for constructing the iTree $\itree_i$. The RIF $\forest$ is then formed by the union of all trees $\itree_1,\cdots, \itree_t$ constructed in this manner.

The scoring subroutine operates in a similar manner. For any given point $p \in \mathbb{R}^d$, the process unfolds as follows: For each tree $\itree_i$, we begin by randomly rotating the point $p$ using the matrix $\rotationmatrix_i$, resulting in the transformed point $p_i$. Subsequently, the scoring function of tree $\itree_i$ is invoked to calculate the score $s_i$. These computed scores $s_1,\cdots, s_t$ are then provided as input to the algorithm $\combinescore(s_1,\cdots,s_t)$ to derive the anomalous score $s(p)$ for the point $p$.


\begin{algorithm}[ht]
 \caption{{\scoring}}
\label{alg:rif:scoring}
 \KwData{RIF forest $\forest = \cup_{i=1}^t (\itree_i, \rotationmatrix_i)$ and an arbitrary point $p$}
 \KwResult{The anomalous score for point $p$ computed using $\forest$}
\For{$i=1$ to $t$}{
	Let $p_i = p \times \rotationmatrix_i$ be the rotation of point $p$ rotation matrix $\rotationmatrix_i$\;
	Use iTree $\itree_i$ to compute anomaly score $s_i$ for point $p_i$\;
}
Invoke Algorithm $\combinescore(s_1,\cdots,s_t)$ to compute anomalous score $s(p)$ of point $p$\;
Return $s(p)$
\end{algorithm}


\subsection{Discussion of issues identified by Hariri et al.~\cite{DBLP:journals/tkde/HaririKB21}}
\label{sec:issues}
As we explained in the introduction, the idea of randomly rotating datasets before applying the iForest construction was also explored in~\cite{DBLP:journals/tkde/HaririKB21}. However, the authors faced various challenges with this approach, leading them to adopt a more sophisticated partitioning method using hyperplanes with random slopes, as proposed in EIF. \cite{DBLP:journals/tkde/HaririKB21}. Below, we explicitly address how the Rotated Isolation Forest (RIF) resolves these challenges.

\paragraph{Rotation in Higher Dimensions}
We begin by addressing the fifth issue raised by Hariri et al.~\cite{DBLP:journals/tkde/HaririKB21}. 
The resolution of this issue provides insights into answering the other questions. 
They mentioned that “\emph{The rotation is not obvious in higher dimensions than 2-D. For each tree we can pick a random axis in the space and perform planar rotation around that axis, but there are many other choices that can be made, which might result in inconsistencies among different runs.}” This is indeed the problem with the way that they do random rotation. In particular, they pick a random axis and then perform a planar random rotation around that axis. This is a naive way of choosing a random rotation and doing it in higher dimensions is not possible. Our random rotation is completely different. We start by generating a random matrix $A$ of size $d\times d$ with elements drawn from a  normal probability distribution with zero mean  and unit variance. Then, we perform the QR decomposition on the random matrix $A$ to obtain an orthogonal matrix $Q$ and an upper triangular matrix $R$, such that $A = QR$. Since $Q$ is orthogonal, its determinant is either $+1$ or $-1$. If it is $+1$, the matrix $Q$ is already a random rotation matrix. If it is $-1$, we can flip the sign  of one of its columns to make the determinant $+1$ and in turn, to obtain the random rotation matrix $Q$. The matrices $A$ and $Q$ are both of size $d\times d$. Thus, it works for any $d$-dimensional Euclidean space $\mathbb{R}^d$.

\paragraph{Tagging Trees with Rotation Information}
In~\cite{DBLP:journals/tkde/HaririKB21}, Hariri et al. wrote that “\emph{Each tree has to be tagged with its unique rotation so that when we are scoring observed data, we can compensate for the rotation in the coordinates of the data point.}” Let us first see what information we need to store for iForest, EIF and the proposed method RIF. The iForest, stores for every node of every tree two numbers that correspond to the random dimension and the random value along that the random dimension. In the EIF, at every node of every tree, they choose a random slope which is a random  normal vector $n$ drawn uniformly over the unit $d$-dimensional sphere and a random intercept $p$ that is drawn from a uniform distribution over the range of values present at that node. For every point $x \in \mathbb{R}^d$, we then compute the test  $(x-p)\cdot n$. If this is negative, $x$ is in one side of the cut, otherwise on the other side. Both $n$ and $p$ are $d$-dimensional vectors. Thus, for every tree the number of values that we keep is the number of nodes per tree multiplied by $2d$ for $d$-dimensional vectors $n$ and $p$. For RIF, besides the values that iForest stores, we store only the random rotation matrix $Q$ what needs $d^2$ values. If the number of nodes in a tree is more than the number of dimensions (which is often the case), the values that we store for RIF are smaller than the $d$-dimensional vectors that we need to store for EIF. 
For a given point $x$, in the RIF, before we feed it into iForest, we multiply $x$ by $Q$. For EIF, at every node we need to compute the test $(x-p)n$. Therefore, tagging a unique rotation for RIF is much simpler than tagging random rotations at every node that is done in EIF.

\paragraph{Rectangular Bias Averaging}
The second issue raised in~\cite{DBLP:journals/tkde/HaririKB21} states: "\emph{Even though the ensemble results seem good, each tree still suffers from the rectangular bias introduced by the underlying algorithm. In a sense, the problem is not resolved but only averaged out.}" This issue does not arise in RIF because the random rotation matrices applied before iForest effectively randomize the orientation of the rectangles used by iForest. As a result, the rectangular regions are no longer axis-aligned, eliminating any bias toward specific axes or fixed directions.

\paragraph{Scalability to Large Datasets and High Dimensions}
 As explained in response to the fifth issue, we utilize QR decomposition to efficiently generate orthogonal random rotation matrices. This approach scales effectively with large datasets and high-dimensional spaces, ensuring that the random rotations remain computationally feasible while preserving the orthogonality necessary for our method. The use of QR decomposition allows us to handle the complexities of high-dimensional data without sacrificing performance.

 \paragraph{Handling Datasets Lacking Symmetries}
 RIF is designed to handle datasets with irregular structures or lacking symmetries. The random rotations introduce diversity in the orientations of the data (See Figure~\ref{fig:rotation:help}). This ensures that the algorithm does not depend on any specific symmetrical properties. By averaging results across multiple random rotations, RIF achieves robustness, even in cases where datasets exhibit asymmetry or irregular patterns.

\paragraph{Extra Bookkeeping and Metadata Storage}


\section{Empirical results}
Next, we present our empirical results. All experiments were conducted using Google Colab and involved both synthetic and real datasets. 
Regarding synthetic datasets, we conducted experiments on two and three-dimensional datasets. Specifically, the two-dimensional datasets consist of: (1) a single spherical Gaussian dataset, (2) two spherical Gaussians, (3) two skewed Gaussians, and (4) a sinusoidal dataset. Additionally, for the three-dimensional dataset, we utilized the Swiss roll dataset.

As for real datasets, we collected nine distinct datasets. Subsequently, we elaborate on these datasets and provide an analysis of the outputs produced by the iForest, EIF, and RIF algorithms applied to these datasets.


\subsection{Two dimensional datasets}
\paragraph{One gaussian distribution}
The initial two-dimensional dataset comprises a cluster of points generated according to a single spherical Gaussian distribution $N(\mu=(0.5,0.5), \sigma=0.07)$. Here, the mean of the distribution is located at $\mu=(0.5,0.5)$ with a standard deviation of $\sigma=0.07$. We generated a total of $2000$ points from this distribution. Two experiments were conducted, involving the addition of anomalies to the underlying space $[0,1]^2$.

\begin{figure}[h]
    \centering
    \includegraphics[width=0.7\linewidth]{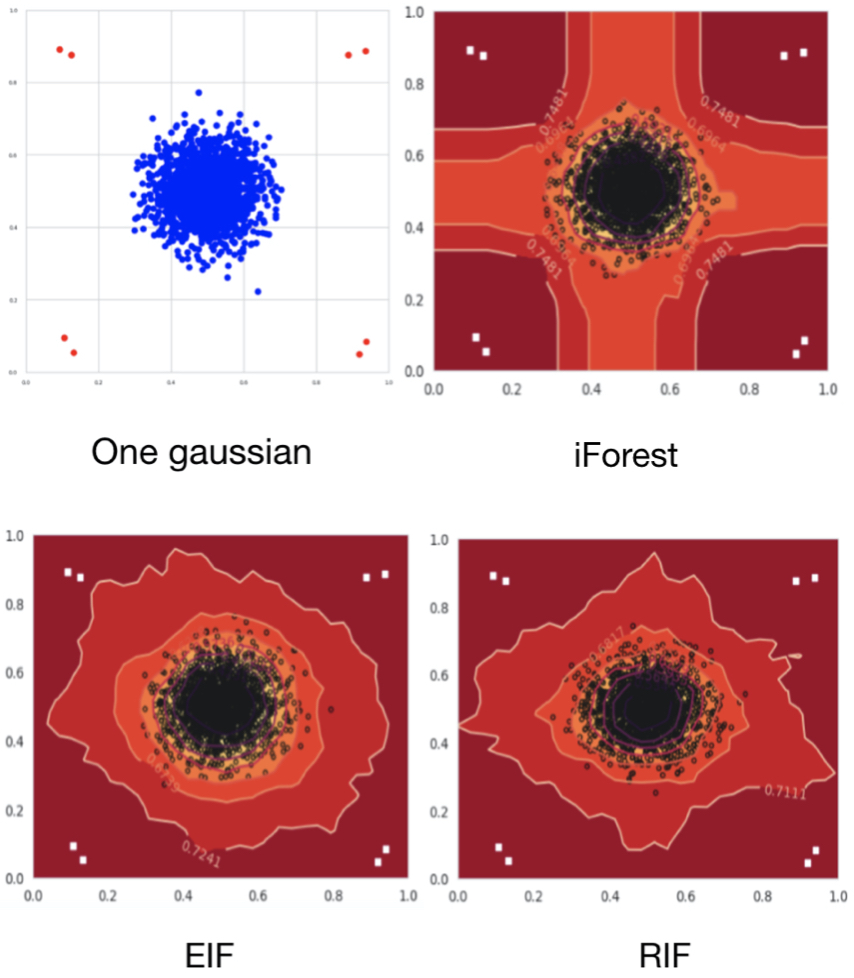}
    \caption{One gaussian distribution with two anomaly points at each corner (top-right, bottom-right, bottom-left, and top-left).
    The AUC score for all methods is $1.0$ as they can detect all anomalies (i.e., white points).}
    \label{fig:ghost:oneblob:corner}
\end{figure}

In the first experiment, we placed $8$ anomaly points at each corner of the underlying space $[0,1]^2$. This arrangement involved positioning two anomaly points at the top-right, two at the bottom-right, two at the bottom-left, and two at the top-left corners. Refer to the left sub-figure in Figure~\ref{fig:ghost:oneblob:corner} to visualize the dataset after adding anomaly points. Subsequently, we employed the construction algorithms of iForest, EIF, and RIF on this dataset. Given the presence of $8$ anomaly points and $2000$ normal points, the contamination ratio of the data was calculated as $\frac{8}{2008} \approx 0.004$.

For each algorithm, we set the number of trees $t=100$ and the sample size $s=256$. Furthermore, to create the heatmap, we divided the underlying space $[0,1]^2$ into a grid consisting of $30$ rows and $30$ columns. This resulted in a total of $900$ grid points. Subsequently, we utilized the scoring function of iForest, EIF, and RIF to compute the anomalous score of these $900$ points based on the forests constructed by each algorithm. The heatmaps depicting these results are presented in Figure~\ref{fig:ghost:oneblob:corner}. 
In each sub-figure, black points denote normal points, while white points represent anomaly points. The color intensity indicates the anomalous score, with darker shades indicating higher anomalous scores. Moreover, the boundaries of areas with anomalous scores are outlined in these sub-figures.


Additionally, we computed the AUC score for these results, which was found to be one. This outcome aligns with our expectations, as we intentionally placed $8$ anomaly points at the corners of the space $[0,1]^2$. Notably, in the sub-figure corresponding to the output of iForest, a sizable area resembling a cross with an orange hue is observed. 
This area, known as the "ghost cluster," is unique to the output of iForest and is not observed in the outputs of EIF and RIF.


\begin{figure}[h]
    \centering
    \includegraphics[width=0.7\linewidth]{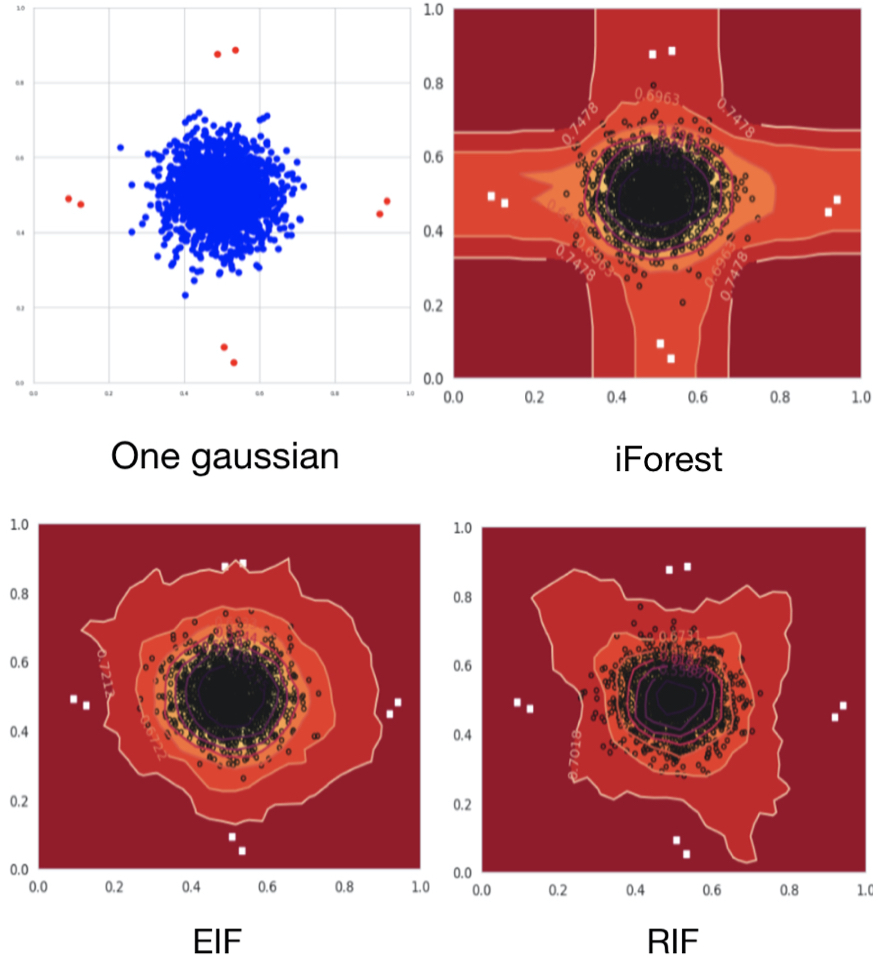}
    \caption{One gaussian distribution with two anomaly points at each position north, south, west, and east.
    The AUC score for iForest is $0.68$, and for EIF and RIF is $1.0$.}
    \label{fig:ghost:oneblob:middles}
\end{figure}


In a second experiment using the same dataset configuration as the first one, we positioned the $8$ anomaly points at different locations—specifically, north, south, west, and east of the dataset. The resulting heatmap is depicted in Figure~\ref{fig:ghost:oneblob:middles}. In this experiment, the AUC score for iForest was $0.68$, while for EIF and RIF, it was $1.0$. Consequently, the conclusion drawn from this experiment is that EIF and RIF outperform iForest.

However, the primary question remains: which algorithm, EIF or RIF, outperforms the other? Based on our observations from the heatmaps, we expect that RIF should outperform EIF, as the lighter areas in RIF are more concentrated compared to those in EIF. 
Our subsequent experiment, focusing on a dataset consisting of two spherical Gaussians, 
confirms this observation and demonstrates that the AUC score of RIF significantly outperforms EIF. 
Interestingly, we show that even iForest outperforms EIF for this dataset.


\paragraph{Two gaussian distributions}
The setup of this experiment follows a similar approach to the first two. We create two clusters of points generated according to two spherical Gaussian distributions: $N_1(\mu=(0.8, 0.2), \sigma=0.06)$ and $N_2(\mu=(0.2, 0.8), \sigma=0.06)$. The distribution from which we draw our samples of $2000$ points is a mixture distribution $N$ with equal weights, where $\alpha_1 = \alpha_2 = \frac{1}{2}$. 

\begin{figure}[h]
    \centering
    \includegraphics[width=0.7\linewidth]{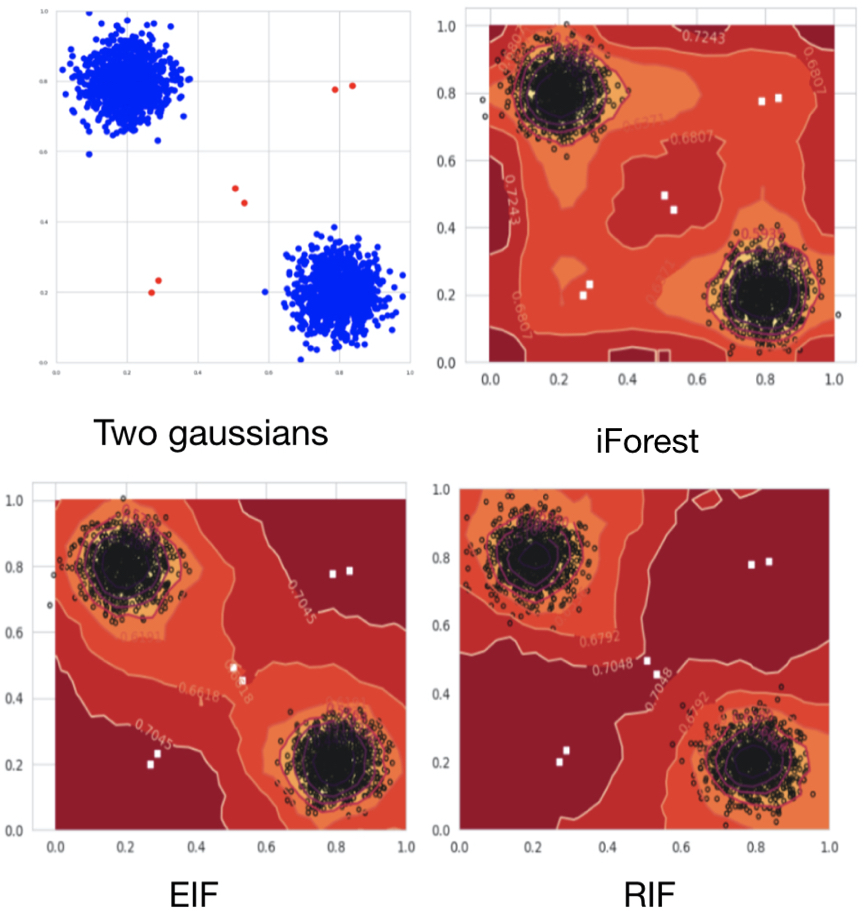}
    \caption{Two gaussian distributions with two anomaly points at each position middle,  top-right, and bottom-left.
    The AUC score for iForest is $0.66$, and for EIF is $0.83$ and for RIF is $1.0$.}
    \label{fig:two:blob:middles:corners}
\end{figure}

To introduce anomalies, we place $6$ anomaly points: two near position $(0.8, 0.8)$ (top-right corner), two near position $(0.25, 0.25)$ (bottom-left corner), and two near the center of the space $[0,1]^2$ at position $(0.5, 0.5)$. These anomaly points are illustrated in white in the left sub-figure of Figure~\ref{fig:two:blob:middles:corners}, while normal points are depicted in blue. With $6$ anomaly points and $2000$ normal points, the contamination ratio of the data is approximately $\frac{6}{2006} \approx 0.003$.


We then apply the construction algorithms of iForest, EIF, and RIF to this dataset and observe an intriguing phenomenon. In the heatmap of iForest, we notice two ghost clusters at the top-right and bottom-left parts. However, in the middle region between the two Gaussian clusters, we observe a dark area indicating a higher anomalous score compared to the ghost clusters. Consequently, the AUC score computed for iForest construction is $0.66$.

In the heatmap generated by EIF, we do not observe the two ghost clusters at the top-right and bottom-left parts. Instead, we observe a ghost cluster in the middle between the two Gaussian clusters. Due to this behavior, the reported AUC score by EIF is $0.88$.

Remarkably, no ghost clusters are observed in the heatmap produced by RIF—neither at the top-right and bottom-left parts nor in the middle. The AUC score returned by RIF is $1.0$, indicating that RIF successfully detects all $6$ anomaly points in this dataset.


\paragraph{Skewed gaussian distributions}
Our third experiment explores Gaussian distributions skewed along a random line. Specifically, we generate two clusters of points according to two spherical Gaussian distributions: $N_1(\mu=(0.2, 0.4), \sigma=0.06)$ and $N_2(\mu=(-0.2, 1), \sigma=0.06)$. These distributions form a mixture distribution $N$ with equal weights, where $\alpha_1 = \alpha_2 = \frac{1}{2}$. 

\begin{figure}[h]
    \centering
    \includegraphics[width=0.7\linewidth]{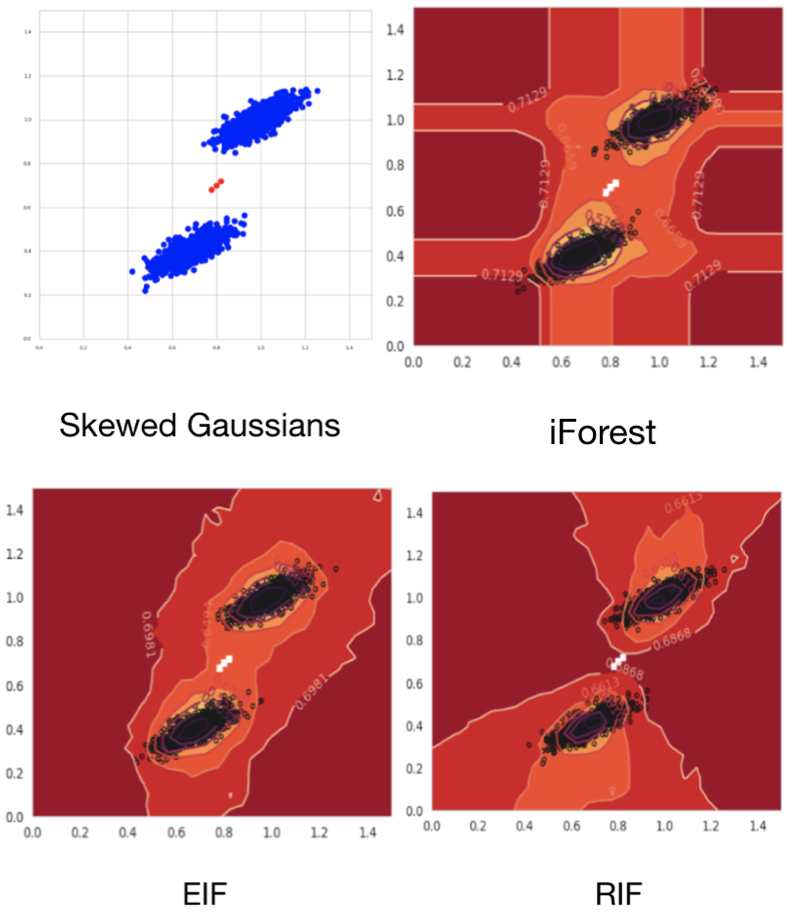}
    \caption{Two skewed gaussian distributions with three anomaly points in the middle. 
    The AUC score for iForest and EIF are $0.50$, and for EIF is $0.83$ and for RIF is $1.0$.}
    \label{fig:two:aniso}
\end{figure}

To introduce skewness, we randomly select an angle and stretch both clusters along that angle. The resulting dataset is illustrated in the left subfigure of Figure~\ref{fig:two:aniso}. Additionally, we include three anomaly points (depicted as white points in the figure) at positions $(0.8, 0.7)$, $(0.82, 0.72)$, and $(0.78, 0.68)$. The contamination ratio of the dataset is approximately $\frac{3}{2003} \approx 0.0015$.


In the heatmap of iForest, we observe strips of ghost clusters—two vertical strips at the top and bottom of both Gaussian clusters, each reasonably thick. Additionally, there are four narrower vertical strips. Moreover, a sizable ghost cluster is visible in the middle between the two Gaussian clusters. The AUC score of iForest is $0.50$, indicating that it fails to detect any of the anomaly points.

Moving to the heatmap of EIF, we do not observe the vertical and horizontal ghost clusters, but a reasonably large ghost cluster remains visible in the middle between the Gaussian clusters. The AUC score of EIF is $0.83$.

However, in the heatmap of RIF, neither the vertical and horizontal ghost clusters nor the central ghost cluster are present. Consequently, RIF reports an AUC score of $1.0$, indicating successful detection of all three anomaly points added to the mixture distribution of Gaussian clusters.

\begin{figure}[h]
    \centering
    \includegraphics[width=0.7\linewidth]{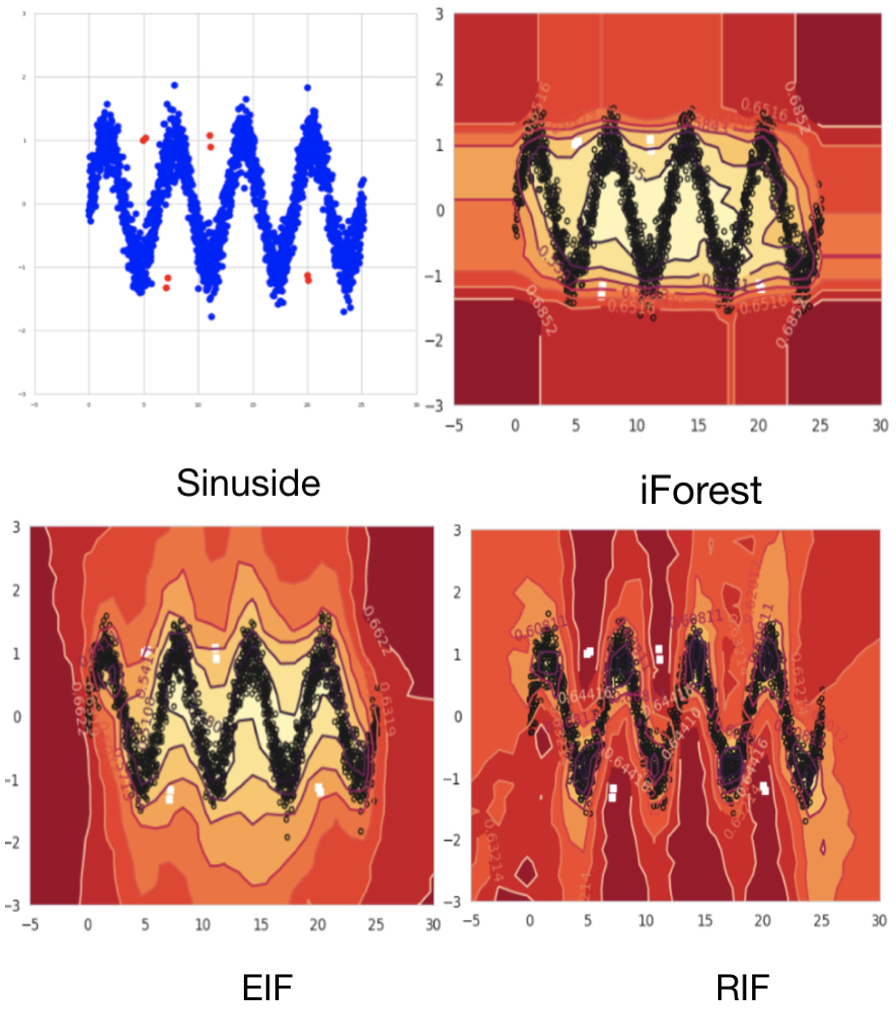}
    \caption{The points are generated according to a Sinuside distribution with white anomaly points are added between peaks and troughs.  
    The AUC score for iForest is $0.50$, and for EIF is $0.83$ and for RIF is $1.0$.}
    \label{fig:sin}
\end{figure}

\paragraph{Sinuside dataset} 
The Sinuside dataset represents our final two-dimensional dataset for experimentation, as depicted in the left subfigure of Figure~\ref{fig:sin}.


In addition to the inherent sinusoidal shape, we intentionally introduced eight anomaly points to this dataset to evaluate the anomaly detection capabilities of different algorithms. Specifically, two anomaly points are situated near $(5, 1)$, another two near $(7, -1)$, two more close to $(10, 1)$, and the final two around $(20, -1)$. These anomaly points are visually distinguished in white in Figure~\ref{fig:sin}.

Notably, neither the heatmap generated by iForest nor the heatmap produced by EIF accurately captures the complex shape of the Sinuside dataset. However, RIF demonstrates superior performance, providing a heatmap that almost entirely illustrates the intricate shape of the Sinuside dataset. Consequently, as expected, the AUC scores of iForest and EIF both amount to $0.5$, indicating their inability to detect any of the anomaly points. Conversely, RIF achieves an AUC score of $1.0$, showcasing its capability to effectively identify all eight anomaly points within the dataset. This marked contrast in performance underscores the effectiveness of RIF in handling datasets with complex geometric structures and anomaly components.


\subsection{Three dimensional datasets}
The \textit{Swiss Roll} dataset is a classic example commonly used in machine learning and data visualization. 
It is designed to illustrate the challenges of learning complex, non-linear manifolds and the effectiveness of algorithms, 
particularly those designed for dimensionality reduction and manifold learning. 
The Swiss Roll dataset can be generated as follows:

\begin{enumerate}
  \item \textbf{2D Grid}: Start with a two-dimensional grid of points. These points are evenly spaced in two dimensions, typically ranging from 0 to 1 in both dimensions.
  \item \textbf{Unrolling}: Roll up the grid in a non-linear fashion to create a three-dimensional spiral. This is done by scaling and bending the grid along one axis while leaving the other axis unchanged.
  \item \textbf{Anomaly}: This could involve randomly perturbing the position of each point slightly to introduce variability and make the dataset more realistic.
\end{enumerate}

The resulting dataset resembles a rolled-up piece of paper or a Swiss roll cake. 
It is characterized by a spiral or coil structure in three dimensions, even though it's embedded in a two-dimensional space. 
The Swiss Roll dataset is particularly useful for testing and demonstrating algorithms that aim to learn the underlying structure of high-dimensional data. 
%
By visualizing the Swiss Roll in two dimensions after applying these techniques, researchers can gain insights into the effectiveness of the algorithms and their ability to capture the intrinsic, non-linear relationships within the data.


\begin{figure}[h]
    \centering
    \includegraphics[width=0.7\linewidth]{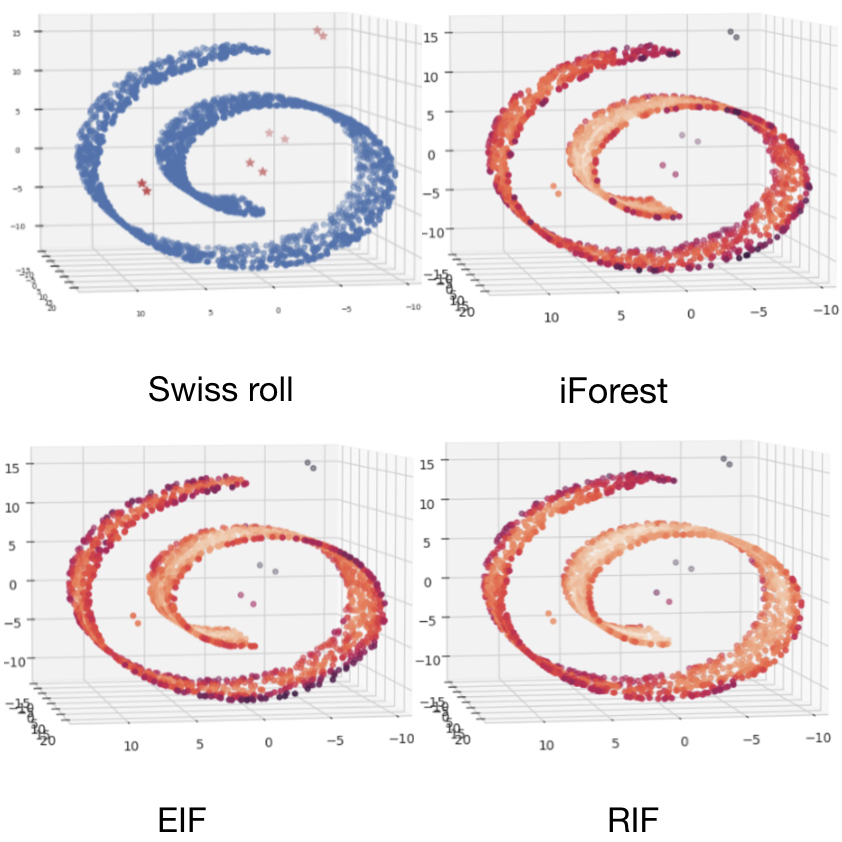}
    \caption{The points are generated according to a Swiss roll distribution with eight anomaly points (red stars) are added. }
    \label{fig:swiss:roll}
\end{figure}


We generated a Swiss Roll dataset, depicted in Figure~\ref{fig:swiss:roll} (left sub-figure), comprising $2000$ points with an added anomaly rate of $0.001$. However, it's important to note that these anomalies are considered part of the normal points. To simulate anomalies detectable by anomaly detection algorithms, we introduced eight additional anomaly points to the dataset, illustrated as red stars in Figure~\ref{fig:swiss:roll}. These anomalies were strategically placed as follows: two points close to $[-5, 0, 0]$, two points near $(-2, 0, -2)$, two points adjacent to $(8, 0, -2)$, and finally, two points neighboring $(-10, -10, 10)$.

Subsequently, we applied the iForest, EIF, and RIF algorithms to this dataset, with the experimental results showcased in Figure~\ref{fig:swiss:roll}. The AUC scores reported by iForest and EIF were $0.68$ and $0.74$, respectively. Consistent with our findings from two-dimensional datasets and real datasets, RIF once again demonstrated superior performance, achieving an impressive AUC score of $0.81$.

%





\subsection{Real datasets}
We conducted a comprehensive array of experiments on real-world datasets, and the details of these datasets along with their properties are meticulously outlined in Table~\ref{tbl:1st}. The table furnishes crucial information such as the dataset names in the first column, followed by the number of instances (examples) and dimensions of each dataset in the second and third columns, respectively. 
The fourth column shows the contamination ratio, which indicates the proportion of instances identified as anomalies.
Finally, the fifth and sixth columns denote the labels assigned to instances corresponding to the normal and anomaly classes.

\begin{center}
\begin{table}[htbp]
\centering 
\begin{tabular*}{\columnwidth}{l|c|c|c|c|c} 
\hline  
Dataset &Size  &Dimension &Contamination  &Normal  &Anomaly\\
\hline   
\hline   
Ionosphere & 351 & 33 & 126 (0.36) & g & B \\
\hline   
Http & 567467 & 3 & 2213 (0.04) & 0 & 1 \\
\hline   
Satellite & 6435 & 36 & 2059 (0.32) & Normal & Anomaly \\
\hline   
Shuttle & 57990 & 9 & 3501 (0.06) & 0 & 1 \\
\hline   
Smtp  & 96554 & 38 & 1183 (0.01) & 0 & 1 \\
\hline   
Cardio  & 1831 & 21 & 190 (0.1) & 0 & 1 \\
\hline   
ForestCover  & 286047 & 11 & 2747 (0.01) & 2 & 4 \\
\hline   
Mammography  & 11183 & 6 & 259 (0.02) & -1 & 1 \\
\hline   
Pima  & 1832 & 21 & 641 (0.35) & 0 & 1 \\
\hline 
\hline 
\end{tabular*}
\caption{The properties of real datasets.} 
\label{tbl:1st}
\end{table}

\end{center}



The outcomes of our experiments involving the iForest, EIF, and RIF algorithms are systematically presented in Table~\ref{tbl:2nd}. For consistency and comparability across experiments, we standardized the parameters by setting the number of trees $t=100$ and the sample size $s=256$ for each algorithm.

To ensure robustness and reliability, we conducted five repetitions of each experiment for every dataset and algorithm combination. The structure of Table~\ref{tbl:2nd} comprises three main columns, each dedicated to iForest, EIF, and RIF, respectively. Within each column, three sub-columns provide insightful metrics: the average AUC score over the five experiment repetitions, the maximum AUC score achieved among these repetitions, and the algorithm contamination set for the respective algorithm.

  \setlength{\tabcolsep}{0.7\tabcolsep}

\begin{table}[!htb]
\centering 
\begin{tabular}{l|ccc|ccc|ccc}

\toprule
Dataset  & \multicolumn{3}{c}{iForest}  & \multicolumn{3}{|c|}{EIF}    & \multicolumn{3}{|c}{RIF}  \\ 
\cmidrule(r){2-4}\cmidrule(r){5-7} \cmidrule(r){8-10}
 & Avg & Max & Cont & Avg & Max & Cont & Avg & Max & Cont\\ \hline

\hline
Ionosphere & 0.807 & 0.817 & 0.25 & 0.821 & 0.829 & 0.25 & \textbf{0.882} & \textbf{0.899} & 0.35 \\
\hline 
Http & 0.893 & 0.902 & 0.20 & 0.917 & 0.952 & 0.10 & \textbf{0.987} & \textbf{0.996} & 0.05 \\
\hline 
Satellite & 0.700 & 0.727 & 0.30 & 0.712 & 0.729 & 0.30 & \textbf{0.731} & \textbf{0.740} & 0.23 \\
\hline 
Shuttle & 0.974 & 0.975 & 0.10 & 0.976 & 0.977 & 0.10 & \textbf{0.983} & \textbf{0.987} & 0.08 \\
\hline 
Smtp & \textbf{0.834} & \textbf{0.842} & 0.05 & 0.813 & 0.825 & 0.05 & 0.830 & 0.825 & 0.05 \\
\hline 
Cardio & 0.846 & 0.880 & 0.25 & 0.849 & 0.872 & 0.30 & \textbf{0.895} & \textbf{0.907} & 0.23 \\
\hline 
ForestCover & 0.810 & 0.806 & 0.20 & \textbf{0.870} & \textbf{0.894} & 0.20 & 0.855 & 0.864 & 0.20 \\
\hline 
Mammography & 0.798 & \textbf{0.820} & 0.20 & 0.796 & 0.810 & 0.25 & \textbf{0.801} & \textbf{0.820} & 0.23 \\
\hline 
Pima & 0.648 & \textbf{0.660} & 0.45 & 0.630 & 0.647 & 0.40 & \textbf{0.653} & 0.655 & 0.40 \\
\bottomrule
\end{tabular}
\caption{The results of real datasets for iForest, EIF, and RIF. 
In this table, "Avg", "Max", and "Cont" are abbreviations for "average", "maximum", and "Contamination", respectively.
} 
\label{tbl:2nd}
\end{table}

It is essential to distinguish between data contamination and algorithm contamination. While data contamination refers to the fraction of instances labeled as anomalies in a dataset, algorithm contamination pertains to the fraction of anomalies designated as input to the algorithm. For instance, when running the iForest algorithm on a dataset $D$ with a contamination rate of $c$, we supply an algorithm contamination $c' \geq c$. Subsequently, the algorithm computes anomalous scores for instances in $D$, sorting them in decreasing order of their scores and identifying anomalies from the top $c'$ fraction of instances in this sorted list. Achieving a good AUC score with an algorithm contamination closer to $c$ implies a lower number of false positives required to detect true anomalies.

Upon careful examination of Table~\ref{tbl:2nd}, it is evident that RIF consistently outperforms the other algorithms across the majority of datasets, exhibiting superior AUC scores while necessitating lower algorithm contamination rates. For example, consider the Http dataset, with a data contamination of $0.04$. While iForest achieves a maximum AUC score of $0.902$ with an algorithm contamination of $0.20$, significantly higher than the dataset's contamination, EIF manages a score of $0.952$ with an algorithm contamination of $0.10$. Remarkably, RIF attains the highest AUC score for the Http dataset, reaching $0.996$ with an algorithm contamination of $0.05$, lower than that of iForest and EIF. Refer to Table~\ref{tbl:2nd} for detailed AUC scores and algorithm contamination metrics for other datasets.

As evidenced by the results obtained from both synthetic datasets and real-world datasets, the RIF algorithm effectively eliminates the fictitious zones created by iForest and EIF, enhancing the consistency of anomaly scores and leading to improved predictions. This is achieved through dynamic representations of input data, increased dataset diversity, and the utilization of oblique hyperplanes.

\subsection{High dimensional real datasets}
Finally, we study the scalability of the RIF algorithm for high-dimensional spaces. 
To this end, we collected nine datasets that are high-dimensional and vary in size, 
with dimensions ranging from  $50$ to $501$ ($[50, 58, 79, 168, 196, 271, 300, 500, 501]$). 
Thus, they are indeed high-dimensional datasets.


\begin{center}
\begin{table}[htbp]
\centering 
\begin{tabular*}{\columnwidth}{l|c|c|c|c|c} 
\hline  
Dataset &Size  &Dimension &Contamination  &Normal  &Anomaly\\
\hline   
\hline   
backdoor & 95329 & 196 & 2330 (0.02) & 0 & 1 \\
\hline   
census & 299285 & 500 & 18569 (0.06) & 0 & 1 \\
\hline   
madelon & 2600 & 501 & 1301 (0.50) & 0 & 1 \\
\hline   
musk & 6598 & 168 & 1018 (0.15) & 1 & 0 \\
\hline   
scene  & 2407 & 300 & 432 (0.18) & 1 & 0 \\
\hline   
Arrhythmia  & 420 & 271 & 208 (0.50) & 1 & 0 \\
\hline   
SpamBase  & 4601 & 58 & 1814 (0.39) & 0 & 1 \\
\hline   
DDos  & 66237 & 79 & 31285 (0.47) & DDoS & BENIGN \\
\hline   
Oil-Spill  & 937 & 50 & 42 (0.04) & -1 & 1 \\
\hline 
\hline 
\end{tabular*}
\caption{The properties of high dimensional real datasets.} 
\label{tbl:3rd}
\end{table}

\end{center}


Table ~\ref{tbl:3rd} provides a summary of the key properties of the datasets, including details such as dimensionality, size, and contamination levels. Meanwhile, Table ~\ref{tbl:4th} presents the results of our experiments, showcasing the performance of iForest, EIF, and RIF. As observed, RIF consistently delivers superior performance compared to iForest and EIF across all these datasets. This consistent performance shows that the capability of the RIF algorithm to effectively handle diverse and challenging data distributions.

\begin{table}[!htb]
\centering 
\begin{tabular}{l|c|c|c|c}

\toprule
Dataset  & {iForest}  & {EIF}    & {RIF}  \\ 
 & Avg & Avg & Avg & Cont\\ \hline

\hline
backdoor & 0.7389 & 0.7478 & \textbf{0.7604} & 0.45 \\
\hline 
census & 0.5835 & 0.5572 & \textbf{0.6357} & 0.45 \\
\hline 
madelon & 0.4980 & 0.5004 & \textbf{0.5113} & 0.10 \\
\hline 
musk & 0.5550 & 0.5560 & \textbf{0.6528} & 0.40 \\
\hline 
scene & 0.5452 & 0.5463 & \textbf{0.6078} & 0.40 \\
\hline 
Arrhythmia & 0.6617 & 0.6662 & \textbf{0.7048} & 0.30 \\
\hline 
SpamBase & 0.5681 & 0.5760 & \textbf{0.6497} & 0.08 \\
\hline 
DDos & 0.5478 & 0.5578 & \textbf{0.5670} & 0.30 \\
\hline 
Oil-Spill & 0.6428 & 0.6786 & \textbf{0.7571} & 0.25 \\
\bottomrule
\end{tabular}
\caption{The results of high dimensional real datasets for iForest, EIF, and RIF. 
In this table, "Avg", and "Cont" are abbreviations for "average", and "Contamination", respectively.
} 
\label{tbl:4th}
\end{table}



%


\section{Conclusion}
This paper introduced Rotated Isolation Forest (RIF) as an enhancement to the popular isolation forest approach for unsupervised anomaly detection. The study addressed limitations observed in the original isolation forest (iForest) and Extended Isolation Forest (EIF), which suffered from axis-parallel and artificial fictitious areas, respectively. By generating dynamic representations of input data and utilizing non-axis-aligned hyperplanes, RIF effectively eliminated these issues, leading to more consistent anomaly scores and improved prediction accuracy. Experimental results on synthetic and real-world datasets provided clear evidence of the superior performance of RIF compared to state-of-the-art  methods.


\begin{thebibliography}{00}
\bibitem{stepanov2021detecting}
  Stepanov, M. D., E. Yu Pavlenko, and Daria S. Lavrova. "Detecting network attacks on software configured networks using the isolating forest algorithm." Automatic Control and Computer Sciences 55.8 (2021): 1039-1050.


\bibitem{arjun2022early}
Arjun, K. P., et al. "Early Prediction of Credit Card Transaction Using Local Outlier Factor and Isolation Forest Tree Machine Learning Algorithms." Applications of Computational Methods in Manufacturing and Product Design: Select Proceedings of IPDIMS 2020. Singapore: Springer Nature Singapore, 2022. 463-472.


\bibitem{tang2022tree}
  Tang, Phat Loi, Thuy-Dung Le Pham, and Tien Ba Dinh. "Tree-Based Credit Card Fraud Detection Using Isolation Forest, Spectral Residual, and Knowledge Graph." International Conference on Machine Learning, Optimization, and Data Science. Cham: Springer Nature Switzerland, 2022.


\bibitem{zhang2022anomaly}
  Zhang, Xiaodong, et al. "Anomaly credit data detection based on enhanced Isolation Forest." The International Journal of Advanced Manufacturing Technology 122.1 (2022): 185-192.

\bibitem{guillardin2023comparing}
  Guillardín, Laura, and John J. MacKay. "Comparing DNA isolation methods for forest trees: quality, plastic footprint, and time-efficiency." Plant Methods 19.1 (2023): 111.


\bibitem{10.1007/978-3-031-27409-1_123}
Rassam, Murad A. "Isolation Forest Based Anomaly Detection Approach for Wireless Body Area Networks." International Conference on Hybrid Intelligent Systems. Cham: Springer Nature Switzerland, 2022.

\bibitem{10.1007/978-981-16-5640-8_3}
Rajeev, Haritha, and Uma Devi. "Detection of credit card fraud using isolation forest algorithm." Pervasive Computing and Social Networking: Proceedings of ICPCSN 2021. Springer Singapore, 2022.


\bibitem{10.1007/978-3-031-26384-2_35}
Ayoub, Mniai, Jebari Khalid, and Pawel Karczmarek. "FUZZY C-MEANS Based Extended Isolation Forest for Anomaly Detection." International Conference on Advanced Intelligent Systems for Sustainable Development. Cham: Springer Nature Switzerland, 2022.

\bibitem{10.1007/978-3-031-57853-3_30}
Downey, Brett E., et al. "Anomaly Detection with Generalized Isolation Forest." International Conference on Advanced Information Networking and Applications. Cham: Springer Nature Switzerland, 2024.

\bibitem{DBLP:journals/tkde/HaririKB21}
Hariri, Sahand, Matias Carrasco Kind, and Robert J. Brunner. "Extended isolation forest." IEEE transactions on knowledge and data engineering 33.4 (2019): 1479-1489.


\bibitem{JMLR:v22:20-600}
Tian, Ye, and Yang Feng. "RaSE: Random subspace ensemble classification." Journal of Machine Learning Research 22.45 (2021): 1-93.



\bibitem{10.1111/rssb.12228}
 Cannings, Timothy I., and Richard J. Samworth. "Random-projection ensemble classification." Journal of the Royal Statistical Society Series B: Statistical Methodology 79.4 (2017): 959-1035.


\bibitem{DBLP:journals/pieee/ApostolidisAMMP21}
Apostolidis, Evlampios, et al. "Video summarization using deep neural networks: A survey." Proceedings of the IEEE 109.11 (2021): 1838-1863.



\bibitem{Mehrotra2017AnomalyDP}
Mehrotra, Kishan G., et al. Anomaly detection. Springer International Publishing, 2017.

\bibitem{DBLP:journals/jmlr/BlaserF16}
Blaser, Rico, and Piotr Fryzlewicz. "Random rotation ensembles." The Journal of Machine Learning Research 17.1 (2016): 126-151.


\bibitem{10072303}
Yeruva, Ajay Reddy, et al. "Anomaly Detection System using ML Classification Algorithm for Network Security." 2022 5th International Conference on Contemporary Computing and Informatics (IC3I). IEEE, 2022.

\bibitem{POURHABIBI2020113303}
Pourhabibi, Tahereh, et al. "Fraud detection: A systematic literature review of graph-based anomaly detection approaches." Decision Support Systems 133 (2020): 113303.


\bibitem{CHOI2022109147}
Choi, Heejeong, et al. "Explainable anomaly detection framework for predictive maintenance in manufacturing systems." Applied Soft Computing 125 (2022): 109147.

\bibitem{PACHAURI2015325}
Pachauri, Girik, and Sandeep Sharma. "Anomaly detection in medical wireless sensor networks using machine learning algorithms." Procedia Computer Science 70 (2015): 325-333.


\bibitem{HILAL2022116429}
Hilal, Waleed, S. Andrew Gadsden, and John Yawney. "Financial fraud: a review of anomaly detection techniques and recent advances." Expert systems With applications 193 (2022): 116429.


\bibitem{Diaconis_Shahshahani_1987}
Diaconis, Persi, and Mehrdad Shahshahani. "The subgroup algorithm for generating uniform random variables." Probability in the engineering and informational sciences 1.1 (1987): 15-32.

\bibitem{DBLP:journals/tkdd/LiuTZ12}
Liu, Fei Tony, Kai Ming Ting, and Zhi-Hua Zhou. "Isolation-based anomaly detection." ACM Transactions on Knowledge Discovery from Data (TKDD) 6.1 (2012): 1-39.


\bibitem{preiss2008data}
Preiss, Bruno R. Data structures and algorithms with object-oriented design patterns in C++. John Wiley and Sons, 2008.


\bibitem{chalapathy2018anomaly}
Chalapathy, Raghavendra, Aditya Krishna Menon, and Sanjay Chawla. "Anomaly detection using one-class neural networks." arXiv preprint arXiv:1802.06360 (2018).


\bibitem{mascaro2014anomaly}
Mascaro, Steven, Ann E. Nicholso, and Kevin B. Korb. "Anomaly detection in vessel tracks using Bayesian networks." International Journal of Approximate Reasoning 55.1 (2014): 84-98.


\bibitem{li2003improving}
Li, Kun-Lun, et al. "Improving one-class SVM for anomaly detection." Proceedings of the 2003 international conference on machine learning and cybernetics (IEEE Cat. No. 03EX693). Vol. 5. IEEE, 2003.


\bibitem{duffield2009rule}
Duffield, Nick, et al. "Rule-based anomaly detection on IP flows." IEEE INFOCOM 2009. IEEE, 2009.


\bibitem{breunig2000lof}
Breunig, Markus M., et al. "LOF: identifying density-based local outliers." Proceedings of the 2000 ACM SIGMOD international conference on Management of data. 2000.


\bibitem{munz2007traffic}
Münz, Gerhard, Sa Li, and Georg Carle. "Traffic anomaly detection using k-means clustering." Gi/itg workshop mmbnet. Vol. 7. No. 9. 2007.


\bibitem{ccelik2011anomaly}
Çelik, Mete, Filiz Dadaşer-Çelik, and Ahmet Şakir Dokuz. "Anomaly detection in temperature data using DBSCAN algorithm." 2011 international symposium on innovations in intelligent systems and applications. IEEE, 2011.


\bibitem{laxhammar2008anomaly}
Laxhammar, Rikard. "Anomaly detection for sea surveillance." 2008 11th international conference on information fusion. IEEE, 2008.


\bibitem{salem2014anomaly}
Salem, Osman, et al. "Anomaly detection in medical wireless sensor networks using SVM and linear regression models." International Journal of E-Health and Medical Communications (IJEHMC) 5.1 (2014): 20-45.


\bibitem{goldstein2012histogram}
Goldstein, Markus, and Andreas Dengel. "Histogram-based outlier score (hbos): A fast unsupervised anomaly detection algorithm." KI-2012: poster and demo track 1 (2012): 59-63.


\bibitem{kwon2005kernel}
Kwon, Heesung, and Nasser M. Nasrabadi. "Kernel RX-algorithm: A nonlinear anomaly detector for hyperspectral imagery." IEEE transactions on Geoscience and Remote Sensing 43.2 (2005): 388-397.


\bibitem{samariya2023comprehensive}
Samariya, Durgesh, and Amit Thakkar. "A comprehensive survey of anomaly detection algorithms." Annals of Data Science 10.3 (2023): 829-850.


\bibitem{fontugne2010mawilab}
Fontugne, Romain, et al. "Mawilab: combining diverse anomaly detectors for automated anomaly labeling and performance benchmarking." Proceedings of the 6th International COnference. 2010.


\bibitem{ding2013compressed}
Ding, Qi, and Eric D. Kolaczyk. "A compressed PCA subspace method for anomaly detection in high-dimensional data." IEEE Transactions on Information Theory 59.11 (2013): 7419-7433.

\bibitem{emmott2013systematic}
Emmott, Andrew F., et al. "Systematic construction of anomaly detection benchmarks from real data." Proceedings of the ACM SIGKDD workshop on outlier detection and description. 2013.


\bibitem{tang2002enhancing}
Tang, Jian, et al. "Enhancing effectiveness of outlier detections for low density patterns." Advances in Knowledge Discovery and Data Mining: 6th Pacific-Asia Conference, PAKDD 2002 Taipei, Taiwan, May 6–8, 2002 Proceedings 6. Springer Berlin Heidelberg, 2002.


\bibitem{papadimitriou2003loci}
Papadimitriou, Spiros, et al. "Loci: Fast outlier detection using the local correlation integral." Proceedings 19th international conference on data engineering (Cat. No. 03CH37405). IEEE, 2003.


\bibitem{jin2006ranking}
Jin, Wen, et al. "Ranking outliers using symmetric neighborhood relationship." Advances in Knowledge Discovery and Data Mining: 10th Pacific-Asia Conference, PAKDD 2006, Singapore, April 9-12, 2006. Proceedings 10. Springer Berlin Heidelberg, 2006.


\bibitem{kriegel2009loop}
Kriegel, Hans-Peter, et al. "LoOP: local outlier probabilities." Proceedings of the 18th ACM conference on Information and knowledge management. 2009.


\bibitem{chandola2009anomaly}
Chandola, Varun, Arindam Banerjee, and Vipin Kumar. "Anomaly detection: A survey." ACM computing surveys (CSUR) 41.3 (2009): 1-58.


\bibitem{pn2005introduction}
Mining, What Is Data. Introduction to data mining. New Jersey: Pearson Education, Inc, 2006.


\bibitem{ester1996density}
Ester, Martin, et al. "A density-based algorithm for discovering clusters in large spatial databases with noise." kdd. Vol. 96. No. 34. 1996.


\bibitem{guha2016robust}
Guha, Sudipto, et al. "Robust random cut forest based anomaly detection on streams." International conference on machine learning. PMLR, 2016.


\bibitem{hariri2018batch}
Hariri, Sahand, and Matias Carrasco Kind. "Batch and online anomaly detection for scientific applications in a Kubernetes environment." Proceedings of the 9th Workshop on Scientific Cloud Computing. 2018.


\bibitem{van2009dimensionality}
Van Der Maaten, Laurens, Eric O. Postma, and H. Jaap Van Den Herik. "Dimensionality reduction: A comparative review." Journal of machine learning research 10.66-71 (2009): 13.


\bibitem{achlioptas2001database}
Achlioptas, Dimitris. "Database-friendly random projections." Proceedings of the twentieth ACM SIGMOD-SIGACT-SIGART symposium on Principles of database systems. 2001.

\bibitem{sorbo2023navigating}
Sørbø, Sondre, and Massimiliano Ruocco. "Navigating the metric maze: A taxonomy of evaluation metrics for anomaly detection in time series." Data Mining and Knowledge Discovery 38.3 (2024): 1027-1068.

\bibitem{li2023explainable}
Li, Zhong, and Matthijs Van Leeuwen. "Explainable contextual anomaly detection using quantile regression forests." Data Mining and Knowledge Discovery 37.6 (2023): 2517-2563.

\bibitem{calikus2022wisdom}
Calikus, Ece, et al. "Wisdom of the contexts: active ensemble learning for contextual anomaly detection." Data Mining and Knowledge Discovery 36.6 (2022): 2410-2458.

\bibitem{zhang2021aurora}
Zhang, Lin, et al. "AURORA: A Unified fRamework fOR Anomaly detection on multivariate time series." Data Mining and Knowledge Discovery 35.5 (2021): 1882-1905.

\bibitem{lu2023damp}
Lu, Yue, et al. "DAMP: accurate time series anomaly detection on trillions of datapoints and ultra-fast arriving data streams." Data Mining and Knowledge Discovery 37.2 (2023): 627-669.

\bibitem{chun2024random}
Chun, Jaewan, et al. "Random walk with restart on hypergraphs: fast computation and an application to anomaly detection." Data Mining and Knowledge Discovery 38.3 (2024): 1222-1257.

\bibitem{biedebach2023anomaly}
Biedebach, Luka, et al. "Anomaly detection in sleep: detecting mouth breathing in children." Data Mining and Knowledge Discovery 38.3 (2024): 976-1005.

\bibitem{konishi2008information}
Konishi, Sadanori, and Genshiro Kitagawa. Information criteria and statistical modeling. Springer Science and Business Media, 2008.

\bibitem{householder1958unitary}
Householder, Alston S. "Unitary triangularization of a nonsymmetric matrix." Journal of the ACM (JACM) 5.4 (1958): 339-342.


\bibitem{waseem2013decomposition}
Waseem, Muhammad, Muhammad Aslam Noor, and Khalida Ianayat Noor. "Decomposition method for solving system of linear equations." Eng. Math. Lett. 2.1 (2013): 34-41.

\bibitem{harville1998matrix}
Harville, David A. "Matrix algebra from a statistician's perspective." (1998): 164-164.


\bibitem{abdi2007eigen}
Abdi, Hervé. "Metric multidimensional scaling (MDS): analyzing distance matrices." Encyclopedia of measurement and statistics (2007): 1-13.


\bibitem{ahmad2017unsupervised}
Ahmad, Subutai, et al. "Unsupervised real-time anomaly detection for streaming data." Neurocomputing 262 (2017): 134-147.

\end{thebibliography}
\end{document}